\documentclass[10pt,twocolumn,letterpaper]{article}
\usepackage[pagenumbers]{cvpr}
\usepackage{multirow,amsfonts,amsthm,microtype}
\usepackage{placeins}
\makeatletter
\renewcommand\paragraph{\@startsection{paragraph}{4}{\z@}{4pt plus 1pt minus 1pt}{-1em}{\normalfont\normalsize\bfseries}}
\makeatother
\definecolor{cvprblue}{rgb}{0.21,0.49,0.74}

\usepackage[pagebackref,breaklinks,colorlinks,allcolors=cvprblue]{hyperref}
\def\confName{CVPR}
\def\confYear{2027}
\newcommand{\cmark}{\checkmark}

\newif\ifshowrevisions
\showrevisionsfalse
\newcommand{\revision}[1]{#1}
\newif\ifshowcvprchanges
\showcvprchangesfalse

\hypersetup{pdftitle={RoboEdit: Turning Human Videos into Scalable Robot Training Data},pdfauthor={Yaowei Guo, Zhiyang Dou, Zeng Tao, Yuxin Jiang, Yunuo Chen, Yuxiang Ma, Yin Yang, Demetri Terzopoulos, Ying Jiang, Chenfanfu Jiang}}
\title{RoboEdit: Turning Human Videos into Scalable Robot Training Data}
\author{\normalfont
\begin{minipage}[t]{\dimexpr\textwidth-2\tabcolsep\relax}
\centering
Yaowei Guo\textsuperscript{1*}, Zhiyang Dou\textsuperscript{2*}, Zeng Tao\textsuperscript{1}, Yuxin Jiang\textsuperscript{1},\\
Yunuo Chen\textsuperscript{1}, Yuxiang Ma\textsuperscript{2}, Yin Yang\textsuperscript{3},\\
Demetri Terzopoulos\textsuperscript{1\textdagger}, Ying Jiang\textsuperscript{4\textdagger}, Chenfanfu Jiang\textsuperscript{1\textdagger}\\[4pt]
\textsuperscript{1}University of California, Los Angeles\quad
\textsuperscript{2}Massachusetts Institute of Technology\\
\textsuperscript{3}University of Utah\quad
\textsuperscript{4}NVIDIA
\\[3pt]{\small\textsuperscript{*}Equal contribution.\quad\textsuperscript{\textdagger}Equal advising.}\end{minipage}}
\makeatletter
\apptocmd{\@maketitle}{\roboeditteaser}{}{\PackageError{RoboEdit}{Unable to attach teaser to title}{}}
\makeatother
\begin{document}
\newcommand{\roboeditteaser}{%
\par\noindent\begin{minipage}{\textwidth}
\centering
\captionsetup{type=figure,hypcap=false}
\includegraphics[width=\linewidth]{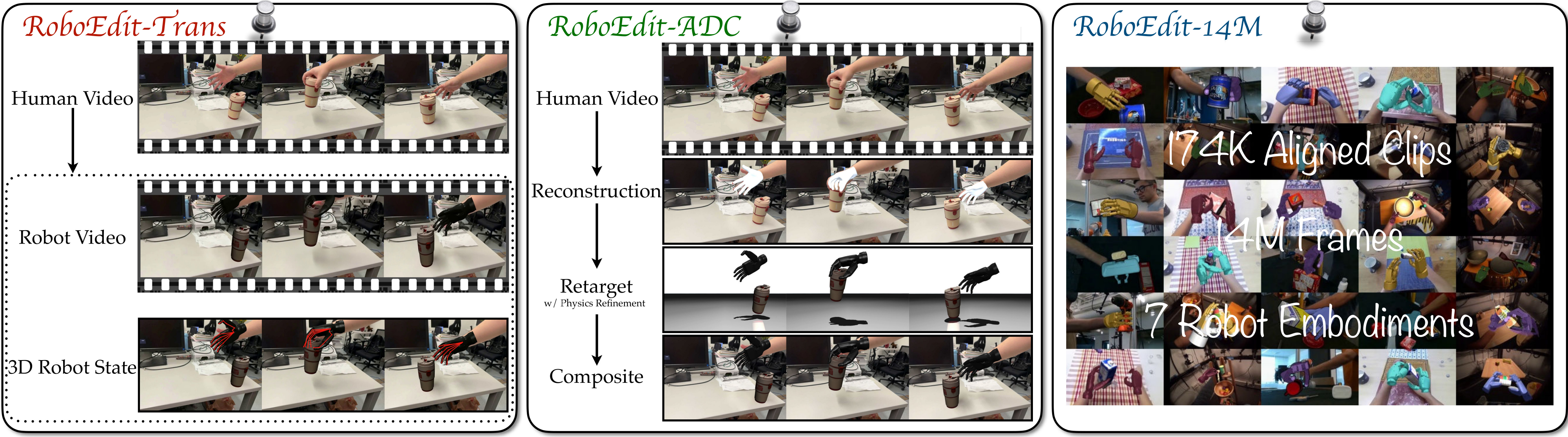}
\caption{RoboEdit Overview: The framework takes an RGB human video and a target robot, to generate a physically plausible robot interaction video and 3D robot-hand states (RoboEdit-Trans). These states enable motion supervision for downstream robot control, while the RoboEdit-ADC pipeline generates the large-scale paired dataset (RoboEdit-14M) required for training.}
\label{fig:teaser}
\end{minipage}\par\vspace{12pt}
}
\maketitle
\begin{abstract}
Collecting robot hand-object interaction data is costly and embodiment-specific, yet abundant human-object videos remain unusable for robot training. We present \emph{RoboEdit}, a human-to-robot video editing suite that transforms human manipulation videos into action-consistent, physically plausible robot videos with aligned 3D hand states. To enable scalable supervision, we introduce \emph{RoboEdit-ADC}, an automatic pipeline that reconstructs and retargets 3D interactions from RGB videos across embodiments. This pipeline generates \emph{RoboEdit-14M}, a large-scale dataset of 174K aligned video pairs (14M frames) spanning seven robot embodiments, diverse scenes, and interaction types. The core editing engine, \emph{RoboEdit-Trans}, employs cross-embodiment adaptation modules to preserve temporal coherence while adapting appearance and motion. It further integrates a 3D Robot-State Decoder to recover per-frame hand states for structured motion supervision. Experiments show that RoboEdit achieves state-of-the-art editing quality and supports downstream robot control policies in real-world manipulation tasks. Ultimately, the RoboEdit suite unlocks the vast potential of unlabeled human videos, providing scalable, high-fidelity visual and 3D motion supervision for generalizable robot learning.

\end{abstract}

\section{Introduction}
\label{sec:introduction}

Video provides rich supervision for robot manipulation by capturing contact, object motion, viewpoint, scene context, and embodiment appearance---details often lost in compact state vectors \citep{black2026pi0visionlanguageactionflowmodel}. However, collecting robot interaction videos is expensive and strictly embodiment-specific: variations in morphology, kinematics, and control prevent direct data transfer across different robot hands \citep{zhang2026unidexrobotfoundationsuite}. Consequently, existing datasets cover only a fraction of the objects, scenes, viewpoints, and behaviors needed for robust, generalizable robot learning.

Human interaction videos offer a scalable alternative, rich with diverse scenes, viewpoints, object motions, and contact-rich behaviors \citep{chao:cvpr2021,engel2023projectarianewtool,liu2024taco}. Yet embodiment mismatches prevent their direct use as robot training data. While prior work has successfully extracted representations or action supervision from human videos \citep{nair2022r3m,wang2023mimicplay}, directly translating these videos into physically plausible target-robot interaction videos remains challenging, particularly in preserving scene context, interaction dynamics, and temporal consistency across robot embodiments \citep{lepert2025phantom}.

This raises a central question: \textit{Can we faithfully transform abundant human videos into scalable, high-fidelity robot interaction videos?}
We answer this question through human-to-robot manipulation video editing, which retargets the interaction to a target robot while preserving the original scene context and temporal dynamics.

We introduce \emph{RoboEdit} (\autoref{fig:teaser}), a suite designed to transform human manipulation videos into scalable, high-fidelity robot training data. It comprises three core components: \emph{RoboEdit-Trans}, a robot-conditioned video editor; \emph{RoboEdit-ADC}, an automatic paired-data curation pipeline; and \emph{RoboEdit-14M}, the resulting large-scale dataset of paired supervision.

\begin{figure*}[!t]
\centering
\includegraphics[width=0.90\textwidth,trim={0 12bp 0 0},clip]{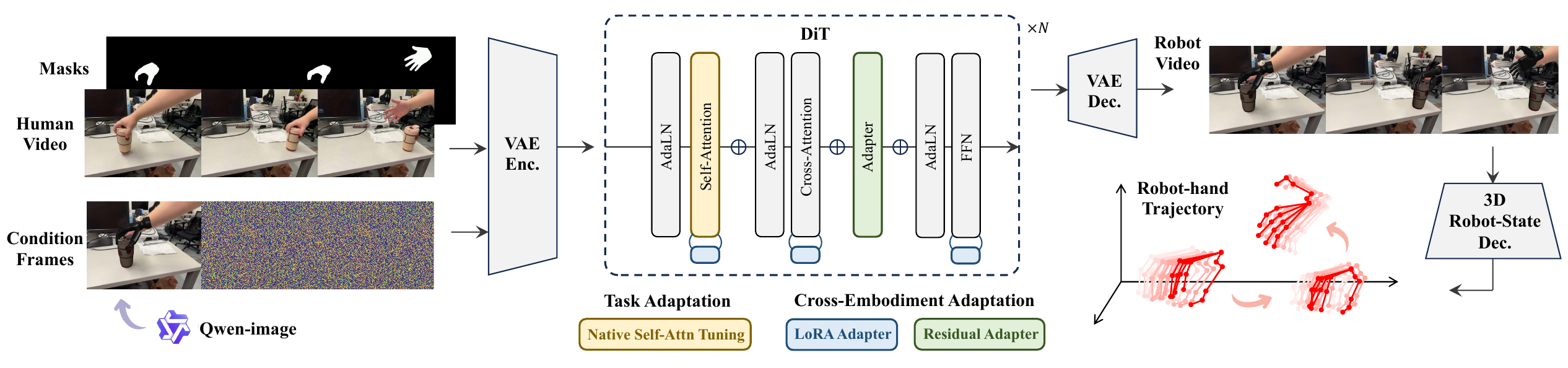}
\caption{RoboEdit-Trans uses LoRA and residual adapter for editing and 3D Robot-State Decoder for state recovery.
}
\label{fig:roboedit-trans}
\end{figure*}

Generating entire robot interactions de novo can disrupt the observed scene context and produce physically implausible motion. RoboEdit-Trans (\autoref{fig:roboedit-trans}) avoids these pitfalls by editing the RGB human interaction video directly, preserving the source scene, camera motion, and object dynamics while retargeting the human embodiment to the target robot.
We introduce cross-embodiment adaptation modules that jointly model cross-embodiment appearance, kinematics, and contact dynamics across diverse robots, ensuring temporal coherence in the generated target-robot video. Furthermore, we integrate a 3D Robot-State Decoder to recover per-frame 3D hand states to serve as structured supervision essential for downstream learning and control.

\begin{figure*}[!t]
\centering
\includegraphics[width=0.85\linewidth]{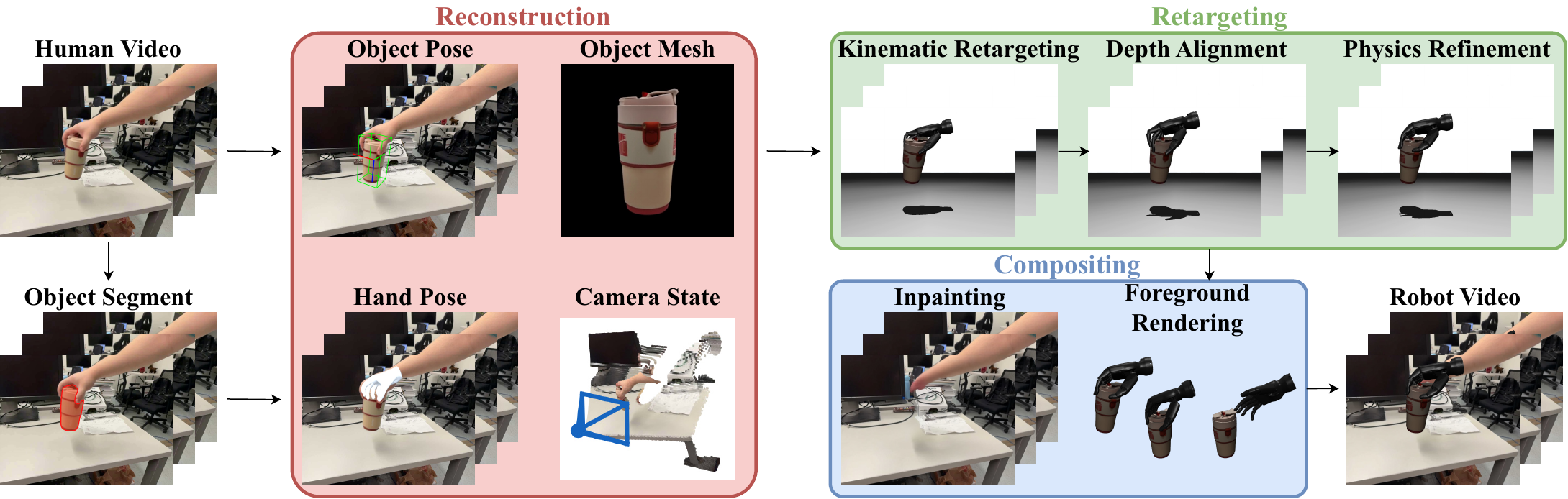}
\caption{RoboEdit-ADC reconstructs 3D interaction from RGB video, performs depth/physics-refined retargeting, and composites the foreground into the inpainted scene.
}
\label{fig:roboedit-adc-pipeline}
\end{figure*}

To provide scalable, aligned supervision, we introduce RoboEdit-ADC (\autoref{fig:roboedit-adc-pipeline}), an automatic pipeline that constructs paired human/robot videos from RGB footage. It first reconstructs the 3D hand-object interaction and camera motion, then retargets the action to a target robot while preserving the original object dynamics and viewpoint. To ensure physical consistency, we incorporate depth regularization and physics-based refinement, effectively reducing artifacts such as object penetration, floating contacts, and temporal jitter. The pipeline concludes by inpainting the human-object regions and compositing the rendered robot interaction into the source scene \citep{ci2025h2r,zi2025minimax_remover}. This process yields RoboEdit-14M (\autoref{fig:dataset-overview}), a massive paired dataset with over 174K aligned video clips (over 14M frames) across 7 robot embodiments, diverse real and synthetic scenes, and varied camera perspectives.

Benchmark comparisons show that RoboEdit-Trans achieves state-of-the-art performance in human-to-robot video editing, outperforming strong baselines in reconstruction fidelity, local editing accuracy, background preservation, and perceptual video quality \citep{he2025openve3mlargescalehighqualitydataset}. Ablation studies confirm consistent gains from each component, while real-world deployments validate the utility of the recovered 3D hand states for downstream control. Collectively, these results demonstrate that human-to-robot video editing transforms abundant, unlabeled human manipulation videos into scalable, high-fidelity robot training data, providing both visual and structured trajectory supervision for robot learning.

Our key contributions are as follows:
\begingroup
\setlength{\topsep}{2pt}
\setlength{\partopsep}{0pt}
\setlength{\itemsep}{2pt}
\setlength{\parsep}{0pt}
\setlength{\parskip}{0pt}
\begin{itemize}
    \item We introduce RoboEdit, an end-to-end human-to-robot video editing suite that transforms RGB human manipulation videos into cross-embodiment robot interaction videos and corresponding structured 3D hand states.
    \item We develop RoboEdit-Trans, a novel video editor featuring robot-aware adaptation modules for cross-embodiment synthesis and a 3D Robot-State Decoder for 3D robot kinematic recovery.
    \item We design RoboEdit-ADC, an automatic data curation pipeline that constructs RoboEdit-14M, a large-scale dataset containing 174K clips (14M frames) across 7 robot embodiments, diverse scenes, and interaction types.
    \item We demonstrate state-of-the-art video editing performance through rigorous benchmarks and ablations, validating the utility of our recovered 3D states for downstream control in real-robot experiments.
\end{itemize}

\endgroup

\section{Related Work}
\label{sec:related-work}

\paragraph{Robot Learning from Human Videos.}
\label{sec:related-human-videos}
Human manipulation videos provide a scalable source of interaction data \citep{grauman2022ego4d,goyal2017somethingsomething}. Prior work leverages this abundance to pretrain transferable visual representations \citep{nair2022r3m,ma2023livlanguageimagerepresentationsrewards}, learn value functions for imitation and reinforcement learning \citep{ma2023vip,zakka2021xirl}, or extract structured interaction priors such as affordances and contact regions \citep{bahl2023affordances,shan2020understanding}.
Closer to our work, recent methods treat human hands as action interfaces or reconstruct and retarget hand-object motion into robot trajectories \citep{shaw2022videodexlearningdexterityinternet,mandikal2022dexviplearningdexterousgrasping,sivakumar2022robotictelekinesislearningrobotic}.
However, these approaches typically convert human videos into intermediate representations or sparse motion supervision. By contrast, RoboEdit directly generates physically-plausible target-robot videos with corresponding 3D hand states, providing rich visual and kinematic supervision in a single framework.

\paragraph{Human-to-Robot Retargeting.}
\label{sec:related-retargeting}

Human-to-robot retargeting adapts human demonstrations to robot embodiments with distinct morphology and kinematics.
Classical approaches solve geometric, task-space, or joint-space objectives to align human wrist, fingertip, grasp, or object-relative motion, forming the basis of teleoperation and data collection \citep{handa2019dexpilotvisionbasedteleoperation,lakshmipathy2024kinematicmotionretargetingcontactrich}.
Alternatively, learning-based methods train cross-embodiment policies for direct imitation \citep{arunachalam2022dexterousimitationeasylearningbased,qin2022dexmvimitationlearningdexterous,shaw2022videodexlearningdexterityinternet}.
For contact-rich manipulation tasks, retargeting further incorporates object dynamics, contact constraints, and physical feasibility through optimization or reinforcement learning \citep{xin2025analyzingkeyobjectiveshumantorobot,pan2025spiderscalablephysicsinformeddexterous}.
Recent monocular pipelines combine 3D hand-object reconstruction with retargeting to derive robot trajectories or policy supervision directly from human videos~\citep{paliwal2026idodexterousmanipulation}. However, these methods typically output sparse trajectories, whereas RoboEdit generates full interaction videos alongside dense 3D states.

\paragraph{Human-to-Robot Video Editing.}

\label{sec:related-video-editing}

Video editing modifies content while preserving temporal coherence. Diffusion-based approaches use inversion~\citep{mokady2023null,qi2023fatezero}, feature manipulation~\citep{geyer2024tokenflow,ku2024anyv2v}, and multimodal conditioning~\citep{liang2025omniv2v}.
Despite advances in video diffusion~\citep{wan2025wan}, most editors assume a fixed embodiment.
Recent work extends video synthesis to robot demonstrations and action learning from human videos~\citep{yang2025xhumanoidrobotizehumanvideos,xie2025human2robotlearningrobotactions}.
Human-to-robot translation must additionally transform morphology and kinematics while preserving interaction semantics.
Mitty~\citep{song2025mitty} and H2R-Grounder~\citep{ci2025h2r} use diffusion models for this translation, while Masquerade~\citep{lepert2026masquerade} composites retargeted robot renderings.
These approaches may produce physically implausible robot--object interactions, including interpenetration and incorrect contacts.
RoboEdit uses explicit 3D penetration and contact checks for trajectory refinement and data filtering, and produces robot videos with articulated 3D trajectories across seven embodiments.

\section{The RoboEdit Suite}
\label{sec:method}

\paragraph{Problem Formulation.}
\label{sec:method-overview}

Given an RGB human hand-object interaction video \(v^{h}_{1:T}\) (where $T$ is the number of frames and $h$ denotes the human domain) and a target robot embodiment \(e\), RoboEdit aims to synthesize a corresponding robot interaction video \(\hat{v}^{r,e}_{1:T}\) in the same scene ($r$ denotes the robot domain) and predict the corresponding robot hand state trajectory \(\hat{q}^{e}_{1:T}\).

Since aligned human/robot video pairs are scarce, we utilize RoboEdit-ADC to construct necessary paired supervision. From the source video, the pipeline reconstructs a 3D representation \(\mathcal{Z}_{1:T}=(H_{1:T},O_{1:T},M_o,m_{1:T},C_{1:T})\), where \(H\) and \(O\) are the human hand and object pose trajectories, \(M_o\) is the object mesh, \(m\) is the edit mask, and \(C\) represents the source camera states. The pipeline then retargets the human trajectory \(H_{1:T}\) relative to the object \(O_{1:T}\) into a robot-hand state trajectory \(q^{e}_{1:T}\), inpaints the human hand and object regions, and renders the robot-object interaction under the original camera motion \(C_{1:T}\) to produce the target video \(v^{r,e}_{1:T}\).

Each curated sample is defined as
\[
\mathcal{D}_i=(v^{h}_{1:T},\,m_{1:T},\,v^{r,e}_{1:T},\,q^{e}_{1:T}).
\]
These samples constitute the RoboEdit-14M dataset (detailed in \autoref{sec:roboedit-dataset}). Finally, RoboEdit-Trans learns a mapping \(F_\theta(v^{h}_{1:T},e)\rightarrow(\hat{v}^{r,e}_{1:T},\hat{q}^{e}_{1:T})\), preserving the source scene, camera motion, and object dynamics while transforming the human interaction to the target embodiment \(e\).

\subsection{RoboEdit-ADC: Automatic Data Curation}
\label{sec:roboedit-adc}
\paragraph{3D Interaction Reconstruction.}
As illustrated in \autoref{fig:roboedit-adc-pipeline}, RoboEdit-ADC reconstructs the full 3D human-object interaction from a single RGB video. We estimate the articulated hand trajectory \(H_{1:T}\) using HaMeR \citep{pavlakos2024reconstructing} and segment the manipulated object using SAM 2 \citep{ravi2024sam2}. The edit masks \(m_{1:T}\) are then derived by combining the object segmentation with the hand mask. For object geometry, we employ TRELLIS \citep{xiang2024structured} to reconstruct the mesh \(M_o\) and FoundationPose \citep{wen2024foundationposeunified6dpose} to track its 6D pose trajectory \(O_{1:T}\). Finally, VGGT \citep{wang2025vggtvisualgeometrygrounded} estimates the source camera state \(C_{1:T}\), including intrinsics, extrinsics, and depth. To ensure physical consistency, we rectify the reconstruction via coordinate-frame and scale calibration before retargeting (see \revision{Appendix}~\ref{sec:supp-camera-depth}).

Monocular HaMeR estimates often suffer from depth and scale ambiguities, such that even an image-aligned hand can be misplaced in 3D space, leading to missed contacts, object penetration, or instability after retargeting.
To address this, we leverage the aligned depth to correct HaMeR's camera-space scale and depth prior to retargeting. We project the recovered palm joints into the depth map and back-project the depth observations into metric 3D palm anchors to estimate a scale factor. We then rescale the camera-space hand reconstruction about the camera center to align its palm depth with the depth observations, while keeping the relative hand articulation fixed. This correction improves the hand-object contact geometry for subsequent retargeting.
Finally, the reconstruction stage outputs the unified hand-object-camera representation:
\[
\mathcal{Z}_{1:T}=\left(H_{1:T},O_{1:T},M_o,m_{1:T},C_{1:T}\right).
\]

\paragraph{Human-Robot Retargeting.}
Given \(\mathcal{Z}_{1:T}\) and a target embodiment \(e\), RoboEdit-ADC retargets the human hand motion using a strategy inspired by SPIDER \citep{pan2025spiderscalablephysicsinformeddexterous}. For anthropomorphic hands, we use SPIDER's wrist-and-fingertip IK in the MuJoCo \citep{todorov2012mujoco} simulator. For morphologically distinct two- and three-finger grippers, we apply specialized  strategies: a two-finger gripper derives its jaw pose and opening from wrist and thumb-index geometry, while a three-finger gripper estimates palm pose and optimizes joint angles to match the thumb, index, and middle fingertips. This yields an initial kinematic trajectory \(q^{e,\text{IK}}_{1:T}\).

To reduce penetration and preserve contact, we jointly refine wrist and finger states while keeping the reconstructed object trajectory fixed. Starting from \(q^{e,\mathrm{IK}}_{1:T}\), we approximately minimize \(\mathcal{L}_{\mathrm{ret}}\) over bounded corrections that respect joint limits and gripper couplings:
\[
\begin{aligned}
\mathcal{L}_{\text{ret}}={}&
\lambda_{\text{track}}\mathcal{L}_{\text{track}}
+\lambda_{\text{geo}}\mathcal{L}_{\text{geo}}\\
&+\lambda_{\text{contact}}\mathcal{L}_{\text{contact}}
+\lambda_{\text{temp}}\mathcal{L}_{\text{temp}}.
\end{aligned}
\]
\(\mathcal{L}_{\mathrm{track}}\) and \(\mathcal{L}_{\mathrm{temp}}\) penalize normalized corrections from the kinematic reference and deviations from neighboring corrections. \(\mathcal{L}_{\mathrm{geo}}\) penalizes robot--object penetration with a 1\,mm clearance margin. We identify source-contact frames when the nearest human fingertip is within 20\,mm of the object. On these frames, \(\mathcal{L}_{\mathrm{contact}}\) penalizes the nearest robot fingertip--object distance only when it exceeds 8\,mm, encouraging contact preservation. We use bounded Jacobian updates and correction smoothing; \revision{Appendix}~\ref{sec:supp-retargeting} details the geometry, losses, and solver.

The optimized trajectory \(q^e_{1:T}\) is then used for rendering the target robot interaction.

\paragraph{Robot Interaction Compositing.}
Given the source video \(v^{h}_{1:T}\), edit masks \(m_{1:T}\), robot and object trajectories \(q^{e}_{1:T}\) and \(O_{1:T}\), and aligned camera states \(C_{1:T}\), RoboEdit-ADC constructs the paired robot video in three steps: (1) MiniMax-Remover \citep{zi2025minimax_remover} inpaints the human hand and object regions within \(m_{1:T}\), producing a clean background \(b_{1:T}\); (2) we render the robot hand and object under \(C_t\) as foreground \(r_t\); and (3) we composite \(r_t\) onto \(b_t\) to obtain the target frame \(v^{r,e}_t\) and the full video \(v^{r,e}_{1:T}\). This process preserves the source background and camera motion while seamlessly substituting the hand-object interaction with the target robot.

\subsection{RoboEdit-Trans: Human-Robot Video Editing}
\label{sec:roboedit-model}

\paragraph{Model Overview and Robot Conditioning.}
While RoboEdit-ADC produces physically plausible robot interaction videos, its multi-stage pipeline is computationally expensive. This motivates RoboEdit-Trans, which aims to generate robot videos more efficiently.
RoboEdit-Trans extends the NovaEdit \citep{pan2026novasparsecontroldense} architecture to enable cross-embodiment robot editing and 3D robot-state prediction (\autoref{fig:roboedit-trans}). We introduce adaptation modules supporting diverse robot embodiments and a 3D Robot-State Decoder to recover the robot hand trajectory. The editor takes a masked human video, providing the scene, camera motion, and object dynamics, and sparse target-robot condition frames specifying the target embodiment and representative hand-object configurations. Let \(z^h\) and \(z^{c,e}\) denote their respective latent representations. The editor predicts target-robot latent \(\hat{z}^{r,e}=G_{\theta}(z^h,z^{c,e})\), where \(G_{\theta}\) is the robot-conditioned editor (\(c\) denotes conditioning) optimized via flow matching.

\begin{figure*}[!t]
\centering
\includegraphics[width=0.90\textwidth]{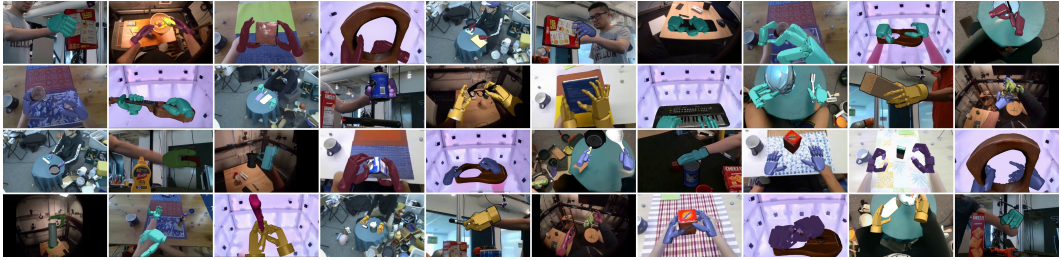}
\caption{RoboEdit-14M spans diverse everyday manipulation tasks across a wide range of robot embodiments.}
\label{fig:dataset-overview}
\end{figure*}

\begin{table*}[!t]
\centering
\caption{Comparison of representative human-to-robot manipulation datasets. Frames follow each work's reported scale; Emb. denotes the number of robot embodiments; Auto denotes curation without per-sample human intervention.}
\label{tab:dataset-comparison}
\normalsize
\small
\setlength{\tabcolsep}{2pt}
\begin{tabular*}{\linewidth}{@{\extracolsep{\fill}}lcccccccc@{}}
\toprule
Dataset & Frames & View & Camera & Scene & Emb. & \shortstack{RGB\\Pair} & \shortstack{Robot\\State} & Auto \\
\midrule
H\&R \citep{xie2025human2robotlearningrobotactions} & \(\sim\)1.2M & Third & Static & Real & 1 & \(\cmark\) & \(\cmark\) & \(\times\) \\
H2R \citep{li2026h2rhumantorobotdataaugmentation} & \(\sim\)1M & Ego & Moving & Real & 3 & \(\cmark\) & \(\times\) & \(\cmark\) \\
UniDex \citep{zhang2026unidexrobotfoundationsuite} & 9M & Ego & Moving & Real & 8 & \(\times\) & \(\cmark\) & \(\times\) \\
X-Humanoid \citep{yang2025xhumanoidrobotizehumanvideos} & 2.8M & Third & Moving & Synth. & 1 & \(\cmark\) & \(\times\) & \(\times\) \\
\textbf{RoboEdit-14M (Ours)} & 14.1M & Ego+Third & Both & Both & 7 & \(\cmark\) & \(\cmark\) & \(\cmark\) \\
\bottomrule
\end{tabular*}
\normalsize

\end{table*}

\paragraph{Cross-Embodiment Adaptation.}
The shared editor learns a common human-to-robot task adaptation, but variations in robot appearance, morphology, kinematics, and contact patterns require cross-embodiment adaptation. We introduce two complementary modules: LoRA adapters~\citep{hu2021loralowrankadaptationlarge} adapt spatiotemporal representations to diverse robot appearance and motion, while residual bottleneck adapters refine embodiment-specific hand geometry and interaction patterns. For input feature \(x\) and intermediate feature \(h\), LoRA and residual-adapter mappings \(f_{\text{LoRA}}\) and \(f_{\text{Ada}}\) are:
\[
\begin{array}{l}
f_{\text{LoRA}}(x)=W_0x+\frac{\alpha}{r}BAx,\\
f_{\text{Ada}}(h)=h+W^{\text{Ada}}_2\phi\!\left(W^{\text{Ada}}_1\text{LayerNorm}(h)\right).
\end{array}
\]
Here, \(W_0\) is the frozen pretrained weight, \(A\) and \(B\) are trainable rank-\(r\) matrices, and \(\alpha\) scales the LoRA update. \(W^{\text{Ada}}_1\) and \(W^{\text{Ada}}_2\) are the trainable down- and up-projection matrices; \(\text{LayerNorm}\) denotes layer normalization, and \(\phi\) is a nonlinear activation. Together, these modules adapt one shared editor across target embodiments while retaining its human-to-robot editing capability.

\paragraph{3D Robot-State Decoder.}
Edited videos do not explicitly encode metric 3D motion, so we leverage a 3D Robot-State Decoder to recover camera-space robot states.
Our 3D Robot-State Decoder first estimates robot states in each frame and then refines them over the full sequence.
For framewise spatial estimation, inspired by heatmap-based 2D hand pose estimation~\citep{iqbal2018handposeestimationlatent}, we use a shared mask-aware image encoder~\citep{he2015deepresiduallearningimage} and design robot-specific prediction heads to estimate per-frame 2D palm anchors and fingertips, palm-frame fingertip coordinates, and joint configurations across diverse embodiments.
A shared camera head predicts camera intrinsics, while the predicted 2D palm anchors are matched with embodiment-specific 3D palm geometry to recover the 3D wrist pose through Perspective-n-Point (PnP).
A temporal Transformer then jointly refines the framewise states over the full sequence to enforce temporal consistency and reduce frame-to-frame jitter, after which forward kinematics produces the complete camera-space robot-hand trajectory.
The decoded states provide structured motion guidance for downstream robot learning and control.

We supervise per-frame 2D localization, 3D geometry, articulation, and camera estimation together with temporal state consistency. More details are in \revision{Appendix}~\ref{sec:supp-decoder}.

\subsection{RoboEdit-14M: Paired Human-Robot Dataset}
\label{sec:roboedit-dataset}

\paragraph{Dataset Overview.}
As shown in \autoref{fig:dataset-overview}, RoboEdit-14M contains 174,547 aligned human/robot video pairs totaling over 14.1M paired frames, constructed from 24,144 human-interaction clips. Each pair includes a target robot video and its retargeted 3D hand trajectory, providing large-scale visual and robot-state supervision.
\autoref{tab:dataset-comparison} shows that RoboEdit-14M uniquely combines automatic curation, paired RGB video, and robot-state labels at 14.1M-frame scale. More details are in \revision{Appendix}~\ref{sec:supp-dataset}.

\paragraph{Source Diversity.}
We source human videos from DexYCB \citep{chao:cvpr2021}, HOT3D \citep{banerjee2024hot3d}, H2O \citep{Kwon_2021_ICCV}, GigaHands \citep{fu2024gigahands}, and TACO \citep{liu2024taco}, covering static and moving egocentric views, single- and bimanual manipulation, and diverse objects and tools. RoboEdit-ADC converts them into 145,459 real-scene pairs spanning 7 embodiments: Inspire, XHand, Ability, SCHUNK SVH, Allegro, Unitree Dex3, and the Franka Panda gripper.

\paragraph{Quality Control.}
To improve and ensure the physical plausibility of the curated training targets, we apply quality control after physics refinement in RoboEdit-ADC. We filter out clips with excessive robot--object penetration or frequent loss of contacts observed in the source human video, retaining only clips that pass these physical checks. This retains 139,058 of the 145,459 real-scene clips (95.60\%). Table~\ref{tab:adc-quality-control} summarizes penetration, contact preservation, and data retention before refinement, after refinement, and after quality control. Further details are provided in \revision{Appendix}~\ref{sec:supp-adc-qc}, ~\ref{sec:supp-dataset}.

\paragraph{Synthetic Scene Augmentation.}
To expand visual diversity, we construct 35,489 synthetic pairs by rendering human and retargeted scene from retained real-scene clips under varied cameras and lighting. We composite both foregrounds onto the same background generated by RoboEngine~\citep{yuan2025roboengineplugandplayrobotdata}, preserving clip alignment while diversifying scene context.

\begin{table*}[!t]
\centering
\caption{Quantitative results on RoboEdit-Trans benchmark. AQ/DD/MS denote VBench Aesthetic Quality/Dynamic Degree/Motion Smoothness; OpenVE is the overall OpenVE-Bench score. \revision{S/M/T/R: single reference / multi-keyframe / text instruction / robot model.}
}
\label{tab:quantitative-comparison}
\small
\setlength{\tabcolsep}{2pt}
\begin{tabular*}{\textwidth}{@{\extracolsep{\fill}}lcccccccccc@{}}
\toprule
\multirow{2}{*}{Method} & \multirow{2}{*}{Params} & \multirow{2}{*}{\revision{Cond.}} & \multicolumn{2}{c}{\revision{Recon.}} & \multicolumn{2}{c}{\revision{Local edit}} & \multicolumn{3}{c}{VBench} & OpenVE \\
\cmidrule(lr){4-5}\cmidrule(lr){6-7}\cmidrule(lr){8-10}\cmidrule(l){11-11}
& & & SSIM $\uparrow$ & LPIPS $\downarrow$ & Edit LPIPS $\downarrow$ & BG SSIM $\uparrow$ & AQ $\uparrow$ & DD $\uparrow$ & MS $\uparrow$ & Overall $\uparrow$ \\
\midrule
VACE & 1.3B & \revision{S} & 0.8764 & 0.1070 & 0.0497 & \textbf{0.9487} & 0.4673 & 0.4867 & 0.9952 & 3.2270 \\
UniVideo & 14B & \revision{S} & 0.3249 & 0.6515 & 0.1043 & 0.4026 & 0.3871 & 0.4133 & 0.9890 & 2.2455 \\
VINO & 13B & \revision{S} & 0.5564 & 0.3348 & 0.0679 & 0.6041 & 0.4488 & 0.4433 & \textbf{0.9967} & \textbf{3.2711} \\
Kiwi-Edit & 5B & \revision{S} & 0.7415 & 0.2059 & 0.0627 & 0.8137 & 0.4307 & 0.4600 & 0.9950 & 3.2386 \\
OmniWeaving & 13B & \revision{S} & 0.8107 & 0.1590 & 0.0625 & 0.8855 & 0.4478 & \textbf{0.6900} & 0.9955 & 3.1634 \\
EditCtrl & 1.3B & \revision{T} & 0.5452 & 0.4295 & 0.0809 & 0.6154 & 0.4319 & 0.4933 & 0.9951 & 3.0319 \\
AnyV2V & 1.3B & \revision{S} & 0.5060 & 0.4837 & 0.0978 & 0.5753 & 0.3891 & 0.1567 & 0.9756 & 2.9023 \\
ReCo & 1.3B & \revision{S} & 0.8433 & 0.1366 & 0.0546 & 0.8869 & \underline{0.4827} & 0.4967 & 0.9939 & 3.1255 \\
\midrule
VACE & 1.3B & \revision{M} & \underline{0.8996} & \underline{0.0778} & \underline{0.0258} & \underline{0.9419} & 0.4768 & 0.5667 & 0.9923 & 3.1446 \\
UniVideo & 14B & \revision{M} & 0.3364 & 0.6464 & 0.1032 & 0.4125 & 0.3836 & 0.4233 & 0.9900 & 2.2089 \\
VINO & 13B & \revision{M} & 0.5539 & 0.3380 & 0.0541 & 0.5897 & 0.4396 & 0.2733 & 0.9944 & 3.1447 \\
\midrule
Mitty & 5B & S & 0.4281 & 0.6251 & 0.0846 & 0.4942 & 0.3699 & 0.4000 & 0.9940 & 2.9892 \\
Masquerade & -- & R & 0.8171 & 0.2046 & 0.0673 & 0.8981 & 0.4770 & 0.6282 & 0.9902 & 2.9495 \\
\textbf{Ours} & 1.3B & \revision{M} & \textbf{0.9282} & \textbf{0.0470} & \textbf{0.0171} & 0.9188 & \textbf{0.4832} & \underline{0.6300} & \underline{0.9956} & \underline{3.2511} \\
\bottomrule
\end{tabular*}
\normalsize

\end{table*}

\section{Experiments}
\label{sec:experiments}

\begin{figure*}[!t]
\centering
\includegraphics[width=0.85\textwidth]{Figures/qualitative_comparison/figure5_qualitative_comparison_cvpr_baselines.pdf}
\caption{Qualitative comparison on four interactions and robot embodiments. The first two columns show human input and RoboEdit-ADC target. Mitty and embodiment-adapted Masquerade occupy the last two columns; all methods use the same cases, timestamps, and crops.
}
\label{fig:qualitative-comparison}

\par\vspace{6pt}

\begin{minipage}[t]{0.58\linewidth}
\captionsetup{type=table}
\vspace{0pt}
\centering
\begin{minipage}[t][24pt][b]{\linewidth}
\captionsetup{justification=raggedright,singlelinecheck=false}
\caption{Ablation of adaptation modules on 300 cases using identical sparse-keyframe conditioning.}
\label{tab:video-editing-ablation}
\end{minipage}\par
\small
\setlength{\tabcolsep}{2pt}
\begin{tabular*}{\linewidth}{@{\extracolsep{\fill}}cccccc@{}}
\toprule
LoRA & Adapter & SSIM $\uparrow$ & LPIPS $\downarrow$ & Edit LPIPS $\downarrow$ & BG SSIM $\uparrow$ \\
\midrule
\(\times\) & \(\times\) & 0.9255 & 0.0494 & 0.0179 & 0.9174 \\
\(\cmark\) & \(\times\) & 0.9264 & 0.0490 & 0.0176 & 0.9181 \\
\(\times\) & \(\cmark\) & 0.9276 & 0.0478 & 0.0173 & 0.9186 \\
\(\cmark\) & \(\cmark\) & \textbf{0.9282} & \textbf{0.0470} & \textbf{0.0171} & \textbf{0.9188} \\
\bottomrule
\end{tabular*}
\normalsize
\end{minipage}\hfill
\begin{minipage}[t]{0.40\linewidth}
\vspace{0pt}
\centering
\begingroup
\ifshowrevisions\color{black}\fi
\begin{minipage}[t][24pt][b]{\linewidth}
\captionsetup{type=table}
\captionsetup{justification=raggedright,singlelinecheck=false}
\ifshowrevisions\captionsetup{labelfont={color=black},textfont={color=black}}\fi
\caption{Decoder accuracy and ablations ($\downarrow$). Shared: shared prediction heads; Temp.: temporal Transformer.}
\label{tab:decoder-quantitative-ablation}
\end{minipage}\par
\small
\setlength{\tabcolsep}{1.5pt}
\renewcommand{\arraystretch}{1}
\begin{tabular*}{\linewidth}{@{\extracolsep{\fill}}lccc@{}}
\toprule
Metric & Shared & w/o Temp. & Full \\
\midrule
Wrist (mm) & 43.15 & 47.21 & \textbf{42.03} \\
Rotation (${}^\circ$) & 25.27 & 33.23 & \textbf{23.28} \\
Fingertip (mm) & 46.27 & 55.85 & \textbf{43.26} \\
Joint (norm.) & 0.2533 & 0.2402 & \textbf{0.2260} \\
\bottomrule
\end{tabular*}
\raggedright
\endgroup
\end{minipage}

\end{figure*}

\subsection{Experimental Setup}
\label{sec:experiments-setup}

\paragraph{Training and Inference Details.}

RoboEdit-Trans builds on NovaEdit~\citep{pan2026novasparsecontroldense} with a Wan2.1-Fun-1.3B-InP backbone~\citep{wan2025wan}. We fine-tuned the backbone on paired human/robot clips for task adaptation, then froze it and trained LoRA and residual adapters jointly for cross-embodiment adaptation. Following NovaEdit, we used 81-frame clips with sparse target-robot keyframes at indices $\{0,10,\ldots,80\}$. At inference, we used Qwen-Image-Edit \citep{wu2025qwenimagetechnicalreport}, fine-tuned on RoboEdit-14M, to generate these keyframes; RoboEdit-Trans then used them to produce the edited video.

We trained the 3D Robot-State Decoder in two stages: the framewise spatial estimator on rendered and composited robot frames, followed by the temporal Transformer on full 81-frame sequences with the spatial estimator frozen. At inference, it recovered the camera-space robot-hand trajectory from the generated video. \revision{Appendix}~\ref{sec:supp-training} provides the details.

\paragraph{Benchmark and Metrics.}

We evaluated RoboEdit-Trans on a fixed benchmark of 300 cases spanning diverse source domains, cameras, interactions, and embodiments.
See Appendix~\ref{sec:supp-benchmark} for benchmark cases distribution.
To evaluate generalization to unseen human demonstrations, we split the dataset by source human video, keeping all robot videos curated from the same human video across different embodiments in the same split.
We report SSIM \citep{1284395} and LPIPS \citep{zhang2018unreasonableeffectivenessdeepfeatures} against the paired reference, edit-region LPIPS against the same reference, and background SSIM against the source outside the edit mask. Additionally, we report VBench \citep{huang2023vbench} measuring visual quality and temporal consistency, and OpenVE-Bench local-replacement protocol \citep{he2025openve3mlargescalehighqualitydataset} assessing prompt compliance, visual naturalness, temporal stability, and physical and motion integrity.

\paragraph{Baselines and Protocols.}

We compare RoboEdit-Trans with VACE \citep{vace}, UniVideo \citep{wei2026univideounifiedunderstandinggeneration}, VINO \citep{chen2026vinounifiedvisualgenerator}, Kiwi-Edit \citep{kiwiedit}, OmniWeaving \citep{pan2026omniweavingunifiedvideogeneration}, EditCtrl \citep{litman2026editctrldisentangledlocalglobal}, AnyV2V \citep{ku2024anyv2v}, and ReCo \citep{zhang2025region}, as well as human-to-robot methods Mitty~\citep{song2025mitty} and Masquerade~\citep{lepert2026masquerade}.
Following each method's supported interface, single-reference methods receive one robot frame. RoboEdit-Trans, VACE, UniVideo, and VINO use the same nine keyframes at indices $\{0,10,\ldots,80\}$ in the multi-reference setting. We extend Masquerade to support all 7 robot hands in our benchmark. More details are in Appendix~\ref{sec:supp-backbone-config}.

\subsection{Qualitative Results}
\label{sec:experiments-qualitative}

\begin{figure*}[!t]
\centering
\includegraphics[width=\linewidth]{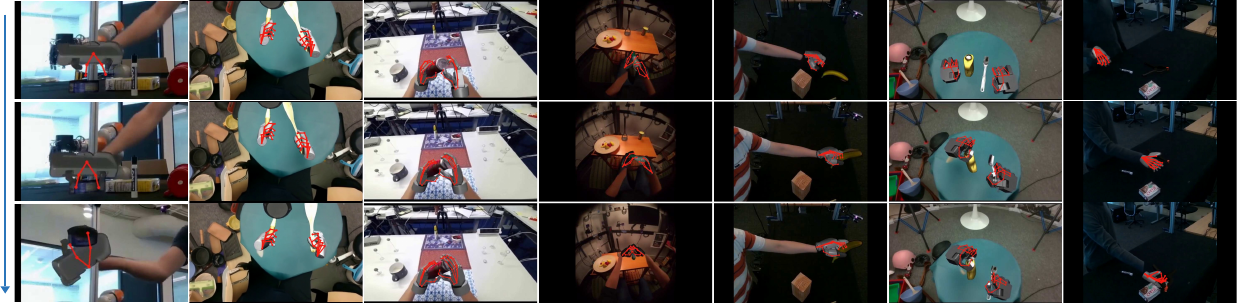}
\caption{3D robot-state predictions on RoboEdit-Trans edited videos across \revision{seven embodiments}.
}
\label{fig:trajectory-decoder-qualitative}

\par\vspace{12pt}

\begin{minipage}[t]{0.48\linewidth}
\centering
\includegraphics[width=0.85\linewidth]{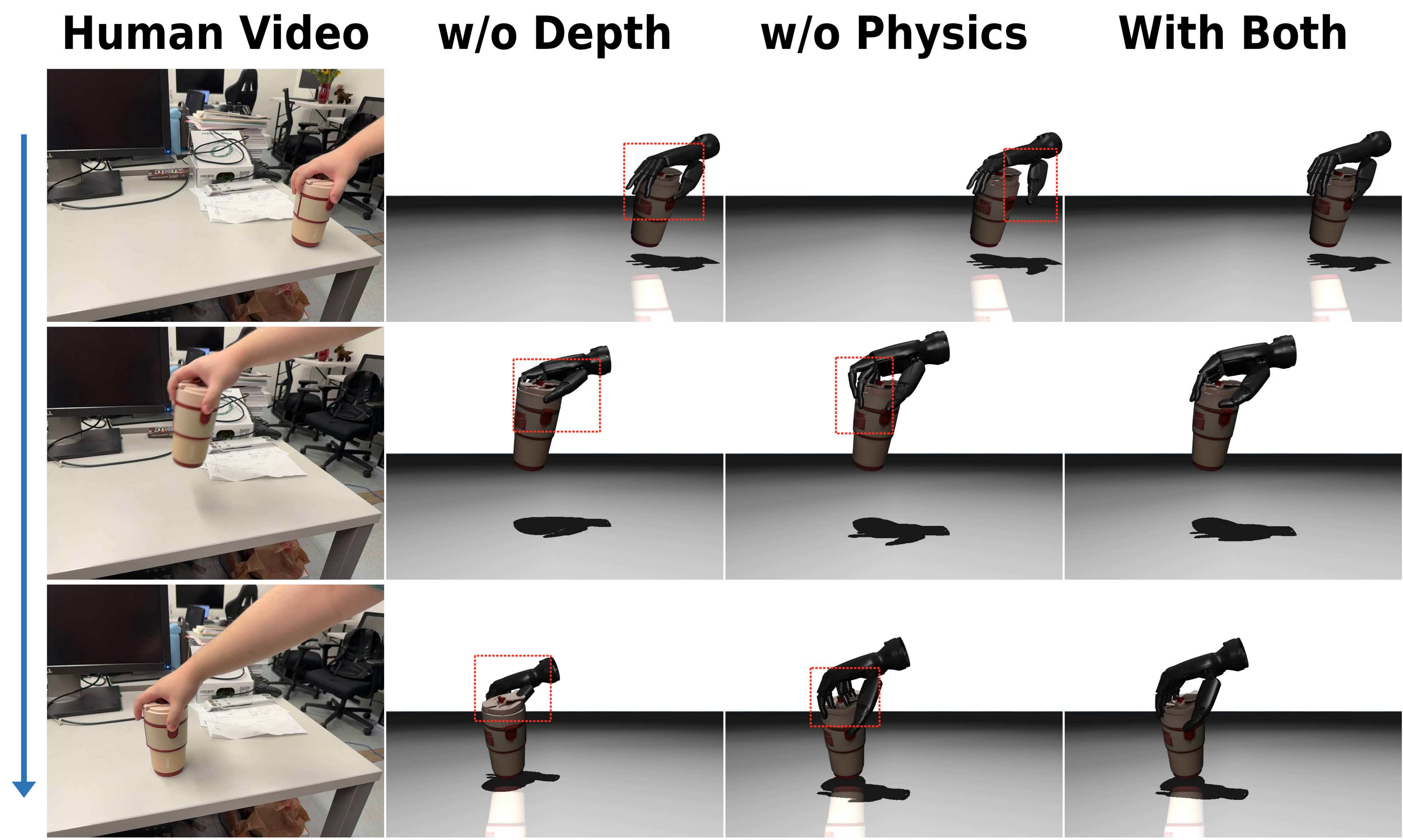}
\caption{Qualitative ablation of RoboEdit-ADC retargeting.
Depth regularization improves hand-object alignment, while physics refinement reduces floating and penetration.}
\label{fig:retargeting-ablation}
\end{minipage}\hfill
\begin{minipage}[t]{0.48\linewidth}
\centering
\includegraphics[width=0.85\linewidth]{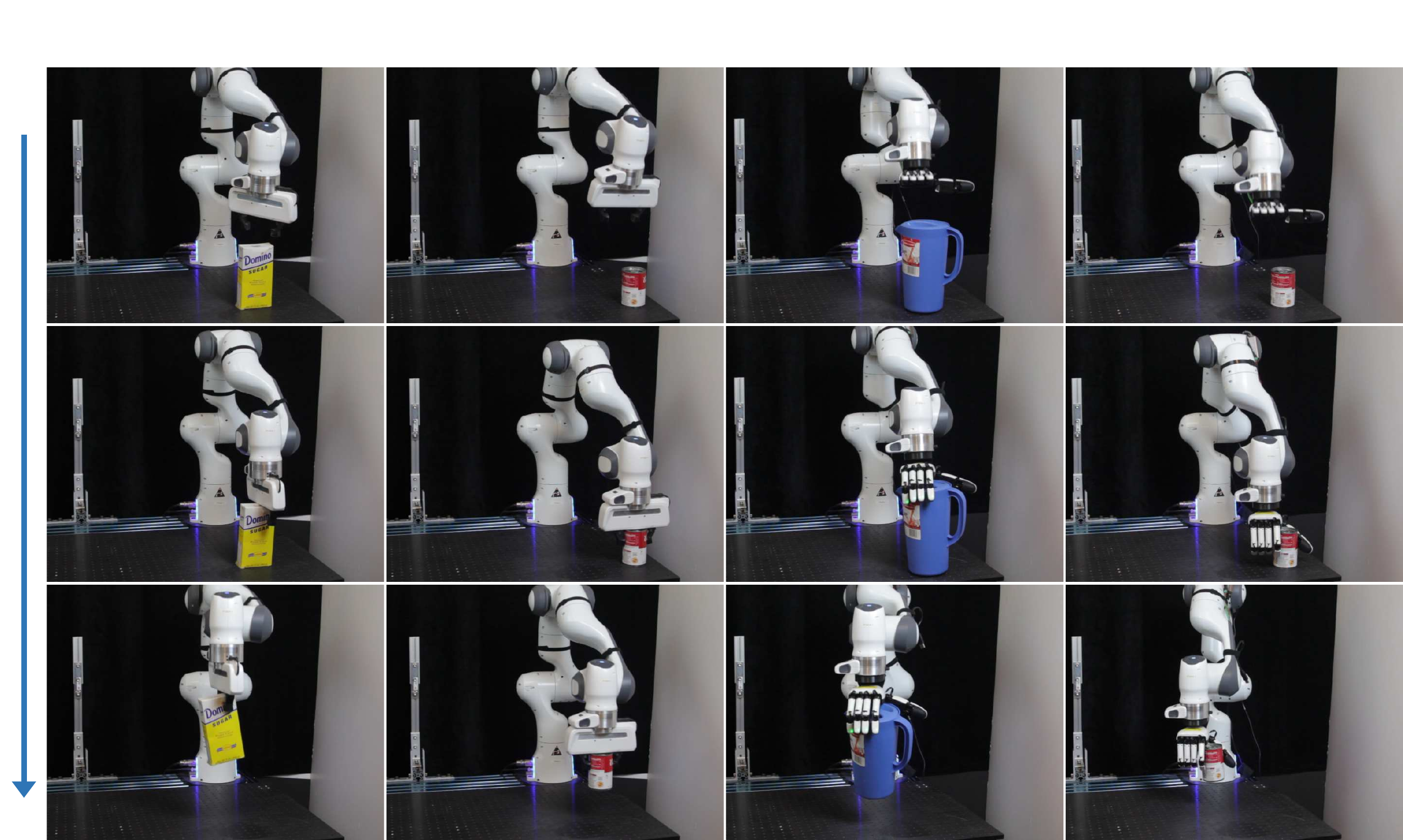}
\caption{Real-robot deployment with 3D Robot-State Decoder trajectories across four YCB-object tasks.
}
\label{fig:real-robot-deployment}
\end{minipage}
\end{figure*}

\autoref{fig:qualitative-comparison} and \autoref{fig:trajectory-decoder-qualitative} show cases from distinct sources and robot hands at non-keyframes. \autoref{fig:qualitative-comparison} compares all methods at matched timestamps and crops. RoboEdit-Trans synthesizes the target embodiment while preserving the scene and object motion, whereas baselines can retain human hands or generate inconsistent hand geometry. \autoref{fig:trajectory-decoder-qualitative} shows that the 3D Robot-State Decoder tracks the generated robot hands and recovers their 3D motion. More results are in Appendix~\ref{sec:supp-qualitative}.

{To evaluate robustness of RoboEdit-ADC and generalization of RoboEdit-Trans, we extend our evaluation to in-the-wild human manipulation videos captured from fixed and egocentric viewpoints. Qualitative results(Appendix~\ref{sec:supp-in-the-wild}) demonstrate that both components can translate these interactions into robot videos across different scenes and views.}

\subsection{Quantitative Comparison}
\label{sec:experiments-quantitative}

\autoref{tab:quantitative-comparison} shows that RoboEdit-Trans achieves the highest reconstruction SSIM and lowest LPIPS and Edit LPIPS among the 300-case methods.
Its slightly lower BG SSIM reflects the differing spatial extent of robot and human hands, which causes valid changes near the edit-mask boundary to count as background errors.

\revision{\autoref{tab:decoder-quantitative-ablation} reports 3D Robot-State Decoder's wrist and fingertip errors in camera space, together with wrist-rotation and normalized joint errors, evaluated on the same 300 cases. Our decoder achieves robot hand estimation errors comparable to those reported for camera-space human hand pose estimation~\citep{ren2026camerahand, handdgp2024}.}

{To complement automatic metrics, we conduct a user study on 30 cases (240 comparisons) sampled from the 300 test cases for overall quality, action preservation, appearance consistency, and contact plausibility. Our method receives 92.5--95.4\% preference. More details are in Appendix~\ref{app:user-study}.}

\textcolor{black}{We also measured runtime for RoboEdit-ADC curation and RoboEdit-Trans inference on the same NVIDIA H100 GPU: RoboEdit-ADC takes approximately 20 minutes across its stages, while RoboEdit-Trans takes approximately 8 minutes including reference keyframe generation, highlighting the efficiency of learned RoboEdit-Trans. A detailed runtime breakdown is in the Appendix~\ref{app:runtime}.}

\subsection{Ablation Study}
\label{sec:experiments-ablation}

\paragraph{Video Editing Components.}

\autoref{tab:video-editing-ablation} reports the contributions of the two cross-embodiment adaptation modules.
Both LoRA and the residual adapter improve all metrics over the backbone-only variant; the residual adapter provides the larger individual gain, while combining both performs best.

\paragraph{\revision{3D Robot-State Decoder.}}

\revision{The variants in \autoref{tab:decoder-quantitative-ablation} replace robot-specific prediction heads with shared heads (Shared) or remove temporal Transformer refinement (w/o Temp.), isolating embodiment-specific prediction and temporal context. Robot-specific heads improve joint estimation, while temporal refinement particularly benefits wrist orientation and fingertip localization.}

\paragraph{Retargeting Quality.}

\autoref{fig:retargeting-ablation} presents the two refinements in RoboEdit-ADC.
Without depth regularization, monocular depth and scale errors propagate to retargeting and displace the robot hand in the scene.
Without physics refinement, kinematic matching produces floating or penetrating fingers.
Table~\ref{tab:adc-quality-control} compares mean penetration and contact-miss before/after physics refinement. More details in \revision{Appendix}~\ref{sec:supp-dataset}.
\begin{table}[!t]
\centering
\caption{{Physical quality and data retention across data curation.}}
\label{tab:adc-quality-control}
\small
\setlength{\tabcolsep}{3pt}
\begin{tabular*}{\linewidth}{@{\extracolsep{\fill}}lccc@{}}
\toprule
Stage & \shortstack{Mean pen.\\(mm) $\downarrow$} & \shortstack{Contact miss\\(\%) $\downarrow$} & \shortstack{Retained\\(\%)} \\
\midrule
Retargeting & {12.20} & {1.439} & {100.00} \\
+ Physics refinement & {2.69} & {0.103} & {100.00} \\
+ Quality control & {2.33} & {0.047} & {95.60} \\
\bottomrule
\end{tabular*}
\end{table}

\subsection{{Simulation and Real-Robot Evaluation}}
\label{sec:experiments-real-robot}

\begin{table}[t]
\centering
\begingroup
\caption{Mean simulation success (\%) over four tasks per hand. }
\label{tab:sim-selected16}
\small
\setlength{\tabcolsep}{3pt}
\begin{tabular*}{\columnwidth}{@{\extracolsep{\fill}}lrr@{}}
\toprule
Embodiment & RoboEdit-ADC & RoboEdit-Trans  \\
\midrule
Allegro & 85.7 & 80.1  \\
XHand & 59.9 & 79.8  \\
Schunk & 75.8 & 55.8  \\
Panda gripper & 84.5 & 65.0  \\
\midrule
\textbf{Mean} & \textbf{76.5} & \textbf{70.2} \\
\bottomrule
\end{tabular*}
\endgroup
\end{table}

{To evaluate our 3D Robot-State Decoder for downstream control, we train residual PPO controllers~\citep{schulman2017proximalpolicyoptimizationalgorithms} in MuJoCo for four YCB-object manipulation tasks each with Allegro, XHand, Schunk, and the Panda gripper. The decoded trajectories specify time-varying target hand poses, guiding the controller's actions while it learns task-specific corrections. Under matched control conditions, RoboEdit-Trans achieves 70.2\% success, compared with 76.5\% for RoboEdit-ADC (Table~\ref{tab:sim-selected16}). Thus, the evaluation shows that RoboEdit-Trans approaches RoboEdit-ADC's control success while reducing end-to-end runtime from about 20 to 8 minutes on the same GPU. More experimental details are in Appendix~\ref{sec:supp-control}.}

We further deploy the trajectory-conditioned controller on a Franka Panda arm equipped with either a Panda gripper or XHand for the four YCB-object tasks (\autoref{fig:real-robot-deployment}). Each task can successfully be completed on the real robot under RoboDojo's protocol~\citep{chen2026robodojo}, demonstrating physical execution with both gripper and dexterous-hand embodiments.

\section{Conclusions}
\label{sec:conclusion}

We introduce RoboEdit, a comprehensive suite that transforms human manipulation videos into physically plausible robot interaction videos and structured 3D hand states. RoboEdit-ADC automatically curates the RoboEdit-14M dataset to provide scalable paired supervision, while RoboEdit-Trans performs cross-embodiment editing and 3D robot state decoding. Our experiments demonstrate state-of-the-art editing quality, consistent gains from each component, and effective motion guidance for downstream control. Collectively, they prove that RoboEdit unlocks the vast potential of abundant human videos, converting them into scalable, high-fidelity robot training data for generalizable learning.

\label{sec:main-end}

\section*{LLM Usage Statement}
\revision{Generative AI tools (ChatGPT) assisted with language editing, literature
lookup, and LaTeX formatting during manuscript preparation. The authors are
responsible for reviewing and verifying all manuscript text, code, figures,
references, and experimental outputs, and take full responsibility for the
final content. Generative models used in the research pipeline, including
synthetic data augmentation, are documented in the method and implementation
details.}
\begingroup
\let\roboeditOriginalBibitem\bibitem
\renewcommand{\bibitem}[2][]{%
  \roboeditOriginalBibitem[#1]{#2}%
  \ifshowrevisions
    \def\roboeditCurrentBibKey{#2}%
    \def\roboeditNewBibKey{lepert2025phantom}%
    \def\roboeditProtocolBibKey{chen2026robodojo}%
    \ifx\roboeditCurrentBibKey\roboeditNewBibKey\color{black}%
    \else\ifx\roboeditCurrentBibKey\roboeditProtocolBibKey\color{black}%
    \else\color{black}\fi\fi
  \fi
  \ignorespaces
}
\small
\bibliographystyle{ieeenat_fullname}
\bibliography{aaai2027}

@inproceedings{lepert2025phantom,
      title={{Phantom}: Training Robots Without Robots Using Only Human Videos},
      author={Lepert, Marion and Fang, Jiaying and Bohg, Jeannette},
      booktitle={Proceedings of The 9th Conference on Robot Learning},
      pages={4545--4565},
      year={2025},
      volume={305},
      series={Proceedings of Machine Learning Research},
      publisher={PMLR},
      url={https://proceedings.mlr.press/v305/lepert25a.html},
}

@misc{black2026pi0visionlanguageactionflowmodel,
      title={$\pi_0$: A Vision-Language-Action Flow Model for General Robot Control},
      author={Kevin Black and Noah Brown and Danny Driess and others},
      year={2026},
      eprint={2410.24164},
      archivePrefix={arXiv},
      primaryClass={cs.LG},
      url={https://arxiv.org/abs/2410.24164},
}

@misc{zhang2026unidexrobotfoundationsuite,
      title={UniDex: A Robot Foundation Suite for Universal Dexterous Hand Control from Egocentric Human Videos},
      author={Gu Zhang and Qicheng Xu and Haozhe Zhang and others},
      year={2026},
      eprint={2603.22264},
      archivePrefix={arXiv},
      primaryClass={cs.RO},
      url={https://arxiv.org/abs/2603.22264},
}

@INPROCEEDINGS{chao:cvpr2021,
  author    = {Yu-Wei Chao and Wei Yang and Yu Xiang and others},
  booktitle = {IEEE/CVF Conference on Computer Vision and Pattern Recognition (CVPR)},
  title     = {{DexYCB}: A Benchmark for Capturing Hand Grasping of Objects},
  year      = {2021},
}

@misc{engel2023projectarianewtool,
      title={Project Aria: A New Tool for Egocentric Multi-Modal AI Research},
      author={Jakob Engel and Kiran Somasundaram and Michael Goesele and others},
      year={2023},
      eprint={2308.13561},
      archivePrefix={arXiv},
      primaryClass={cs.HC},
      url={https://arxiv.org/abs/2308.13561},
}

@article{liu2024taco,
      title={TACO: Benchmarking Generalizable Bimanual Tool-ACtion-Object Understanding},
      author={Liu, Yun and Yang, Haolin and Si, Xu and others},
      journal={arXiv preprint arXiv:2401.08399},
      year={2024}
    }

@misc{banerjee2024hot3d,
      title={{HOT3D}: Hand and Object Tracking in 3D from Egocentric Multi-View Videos},
      author={Prithviraj Banerjee and Sindi Shkodrani and Pierre Moulon and others},
      year={2024},
      eprint={2411.19167},
      archivePrefix={arXiv},
      primaryClass={cs.CV},
      url={https://arxiv.org/abs/2411.19167},
}

@InProceedings{Kwon_2021_ICCV,
      author = {Kwon, Taein and Tekin, Bugra and St\"uhmer, Jan and Bogo, Federica and Pollefeys, Marc},
      title = {H2O: Two Hands Manipulating Objects for First Person Interaction Recognition},
      booktitle = {Proceedings of the IEEE/CVF International Conference on Computer Vision (ICCV)},
      month = {October},
      year = {2021},
      pages = {10138-10148}
}

@misc{fu2024gigahands,
      title={GigaHands: A Massive Annotated Dataset of Bimanual Hand Activities},
      author={Rao Fu and Dingxi Zhang and Alex Jiang and others},
      year={2024},
      eprint={2412.04244},
      archivePrefix={arXiv},
      primaryClass={cs.CV},
      url={https://arxiv.org/abs/2412.04244},
}

@misc{li2026h2rhumantorobotdataaugmentation,
      title={H2R: A Human-to-Robot Data Augmentation for Robot Pre-training from Videos}, 
      author={Guangrun Li and Yaoxu Lyu and Zhuoyang Liu and Chengkai Hou and Jieyu Zhang and Shanghang Zhang},
      year={2026},
      eprint={2505.11920},
      archivePrefix={arXiv},
      primaryClass={cs.RO},
      url={https://arxiv.org/abs/2505.11920}, 
}

@misc{pan2025spiderscalablephysicsinformeddexterous,
            title={SPIDER: Scalable Physics-Informed Dexterous Retargeting},
            author={Chaoyi Pan and Changhao Wang and Haozhi Qi and Zixi Liu and Homanga Bharadhwaj and Akash Sharma and Tingfan Wu and Guanya Shi and Jitendra Malik and Francois Hogan},
            year={2025},
            eprint={2511.09484},
            archivePrefix={arXiv},
            primaryClass={cs.RO},
            url={https://arxiv.org/abs/2511.09484},
}

@misc{pan2026novasparsecontroldense,
      title={NOVA: Sparse Control, Dense Synthesis for Pair-Free Video Editing},
      author={Tianlin Pan and Jiayi Dai and Chenpu Yuan and Zhengyao Lv and Binxin Yang and Hubery Yin and Chen Li and Jing Lyu and Caifeng Shan and Chenyang Si},
      year={2026},
      eprint={2603.02802},
      archivePrefix={arXiv},
      primaryClass={cs.CV},
      url={https://arxiv.org/abs/2603.02802},
}

@inproceedings{vace,
    title = {VACE: All-in-One Video Creation and Editing},
    author = {Jiang, Zeyinzi and Han, Zhen and Mao, Chaojie and Zhang, Jingfeng and Pan, Yulin and Liu, Yu},
    booktitle = {Proceedings of the IEEE/CVF International Conference on Computer Vision},
    pages = {17191-17202},
    year = {2025}
}

@misc{wei2026univideounifiedunderstandinggeneration,
      title={UniVideo: Unified Understanding, Generation, and Editing for Videos},
      author={Cong Wei and Quande Liu and Zixuan Ye and Qiulin Wang and Xintao Wang and Pengfei Wan and Kun Gai and Wenhu Chen},
      year={2026},
      eprint={2510.08377},
      archivePrefix={arXiv},
      primaryClass={cs.CV},
      url={https://arxiv.org/abs/2510.08377},
}

@article{zi2025minimax_remover,
      title={MiniMax-Remover: Taming Bad Noise Helps Video Object Removal},
      author={Bojia Zi and Weixuan Peng and Xianbiao Qi and Jianan Wang and Shihao Zhao and Rong Xiao and Kam-Fai Wong},
      journal={arXiv preprint arXiv:2505.24873},
      year={2025},
}

@InProceedings{huang2023vbench,
     title={{VBench}: Comprehensive Benchmark Suite for Video Generative Models},
     author={Huang, Ziqi and He, Yinan and Yu, Jiashuo and others},
     booktitle={Proceedings of the IEEE/CVF Conference on Computer Vision and Pattern Recognition},
     year={2024}
 }

@misc{he2025openve3mlargescalehighqualitydataset,
      title={OpenVE-3M: A Large-Scale High-Quality Dataset for Instruction-Guided Video Editing},
      author={Haoyang He and Jie Wang and Jiangning Zhang and others},
      year={2025},
      eprint={2512.07826},
      archivePrefix={arXiv},
      primaryClass={cs.CV},
      url={https://arxiv.org/abs/2512.07826},
}

@misc{handa2019dexpilotvisionbasedteleoperation,
      title={DexPilot: Vision Based Teleoperation of Dexterous Robotic Hand-Arm System}, 
      author={Ankur Handa and Karl Van Wyk and Wei Yang and Jacky Liang and Yu-Wei Chao and Qian Wan and Stan Birchfield and Nathan Ratliff and Dieter Fox},
      year={2019},
      eprint={1910.03135},
      archivePrefix={arXiv},
      primaryClass={cs.CV},
      url={https://arxiv.org/abs/1910.03135}, 
}

@misc{arunachalam2022dexterousimitationeasylearningbased,
      title={Dexterous Imitation Made Easy: A Learning-Based Framework for Efficient Dexterous Manipulation}, 
      author={Sridhar Pandian Arunachalam and Sneha Silwal and Ben Evans and Lerrel Pinto},
      year={2022},
      eprint={2203.13251},
      archivePrefix={arXiv},
      primaryClass={cs.RO},
      url={https://arxiv.org/abs/2203.13251}, 
}

@misc{qin2022dexmvimitationlearningdexterous,
      title={DexMV: Imitation Learning for Dexterous Manipulation from Human Videos}, 
      author={Yuzhe Qin and Yueh-Hua Wu and Shaowei Liu and Hanwen Jiang and Ruihan Yang and Yang Fu and Xiaolong Wang},
      year={2022},
      eprint={2108.05877},
      archivePrefix={arXiv},
      primaryClass={cs.LG},
      url={https://arxiv.org/abs/2108.05877}, 
}

@misc{shaw2022videodexlearningdexterityinternet,
      title={VideoDex: Learning Dexterity from Internet Videos}, 
      author={Kenneth Shaw and Shikhar Bahl and Deepak Pathak},
      year={2022},
      eprint={2212.04498},
      archivePrefix={arXiv},
      primaryClass={cs.RO},
      url={https://arxiv.org/abs/2212.04498}, 
}

@misc{lakshmipathy2024kinematicmotionretargetingcontactrich,
      title={Kinematic Motion Retargeting for Contact-Rich Anthropomorphic Manipulations}, 
      author={Arjun S. Lakshmipathy and Jessica K. Hodgins and Nancy S. Pollard},
      year={2024},
      eprint={2402.04820},
      archivePrefix={arXiv},
      primaryClass={cs.GR},
      url={https://arxiv.org/abs/2402.04820}, 
}

@misc{xin2025analyzingkeyobjectiveshumantorobot,
      title={Analyzing Key Objectives in Human-to-Robot Retargeting for Dexterous Manipulation}, 
      author={Chendong Xin and Mingrui Yu and Yongpeng Jiang and Zhefeng Zhang and Xiang Li},
      year={2025},
      eprint={2506.09384},
      archivePrefix={arXiv},
      primaryClass={cs.RO},
      url={https://arxiv.org/abs/2506.09384}, 
}

@misc{sivakumar2022robotictelekinesislearningrobotic,
      title={Robotic Telekinesis: Learning a Robotic Hand Imitator by Watching Humans on Youtube}, 
      author={Aravind Sivakumar and Kenneth Shaw and Deepak Pathak},
      year={2022},
      eprint={2202.10448},
      archivePrefix={arXiv},
      primaryClass={cs.RO},
      url={https://arxiv.org/abs/2202.10448}, 
}

@misc{mandikal2022dexviplearningdexterousgrasping,
      title={DexVIP: Learning Dexterous Grasping with Human Hand Pose Priors from Video}, 
      author={Priyanka Mandikal and Kristen Grauman},
      year={2022},
      eprint={2202.00164},
      archivePrefix={arXiv},
      primaryClass={cs.RO},
      url={https://arxiv.org/abs/2202.00164}, 
}

@misc{paliwal2026idodexterousmanipulation,
      title={Do as I Do: Dexterous Manipulation Data from Everyday Human Videos}, 
      author={Bhawna Paliwal and Haritheja Etukuru and William Liang and Pieter Abbeel and Nur Muhammad Mahi Shafiullah and Jitendra Malik},
      year={2026},
      eprint={2606.19333},
      archivePrefix={arXiv},
      primaryClass={cs.RO},
      url={https://arxiv.org/abs/2606.19333}, 
}

@inproceedings{pavlakos2024reconstructing,
  title={Reconstructing Hands in 3{D} with Transformers},
  author={Pavlakos, Georgios and Shan, Dandan and Radosavovic, Ilija and Kanazawa, Angjoo and Fouhey, David and Malik, Jitendra},
  booktitle={CVPR},
  year={2024}
}

@misc{wen2024foundationposeunified6dpose,
      title={FoundationPose: Unified 6D Pose Estimation and Tracking of Novel Objects}, 
      author={Bowen Wen and Wei Yang and Jan Kautz and Stan Birchfield},
      year={2024},
      eprint={2312.08344},
      archivePrefix={arXiv},
      primaryClass={cs.CV},
      url={https://arxiv.org/abs/2312.08344}, 
}

@article{xiang2024structured,
    title   = {Structured 3D Latents for Scalable and Versatile 3D Generation},
    author  = {Xiang, Jianfeng and Lv, Zelong and Xu, Sicheng and Deng, Yu and Wang, Ruicheng and 
               Zhang, Bowen and Chen, Dong and Tong, Xin and Yang, Jiaolong},
    journal = {arXiv preprint arXiv:2412.01506},
    year    = {2024}
}

@misc{wang2025vggtvisualgeometrygrounded,
      title={VGGT: Visual Geometry Grounded Transformer}, 
      author={Jianyuan Wang and Minghao Chen and Nikita Karaev and Andrea Vedaldi and Christian Rupprecht and David Novotny},
      year={2025},
      eprint={2503.11651},
      archivePrefix={arXiv},
      primaryClass={cs.CV},
      url={https://arxiv.org/abs/2503.11651}, 
}

@inproceedings{mokady2023null,
  title={Null-text inversion for editing real images using guided diffusion models},
  author={Mokady, Ron and Hertz, Amir and Aberman, Kfir and Pritch, Yael and Cohen-Or, Daniel},
  booktitle={Proceedings of the IEEE/CVF conference on computer vision and pattern recognition},
  pages={6038--6047},
  year={2023}
}

@inproceedings{qi2023fatezero,
  title={Fatezero: Fusing attentions for zero-shot text-based video editing},
  author={Qi, Chenyang and Cun, Xiaodong and Zhang, Yong and Lei, Chenyang and Wang, Xintao and Shan, Ying and Chen, Qifeng},
  booktitle={Proceedings of the IEEE/CVF International Conference on Computer Vision},
  pages={15932--15942},
  year={2023}
}

@inproceedings{geyer2024tokenflow,
  title={Tokenflow: Consistent diffusion features for consistent video editing},
  author={Geyer, Michal and Bar Tal, Omer and Bagon, Shai and Dekel, Tali},
  booktitle={International Conference on Learning Representations},
  volume={2024},
  pages={1608--1620},
  year={2024}
}

@article{liang2025omniv2v,
  title={Omniv2v: Versatile video generation and editing via dynamic content manipulation},
  author={Liang, Sen and Yu, Zhentao and Zhou, Zhengguang and Hu, Teng and Wang, Hongmei and Chen, Yi and Lin, Qin and Zhou, Yuan and Li, Xin and Lu, Qinglin and others},
  journal={arXiv preprint arXiv:2506.01801},
  year={2025}
}

@article{wan2025wan,
  title={Wan: Open and advanced large-scale video generative models},
  author={Wan, Team and Wang, Ang and Ai, Baole and Wen, Bin and Mao, Chaojie and Xie, Chen-Wei and Chen, Di and Yu, Feiwu and Zhao, Haiming and Yang, Jianxiao and others},
  journal={arXiv preprint arXiv:2503.20314},
  year={2025}
}

@misc{hu2021loralowrankadaptationlarge,
      title={LoRA: Low-Rank Adaptation of Large Language Models},
      author={Edward J. Hu and Yelong Shen and Phillip Wallis and Zeyuan Allen-Zhu and Yuanzhi Li and Shean Wang and Lu Wang and Weizhu Chen},
      year={2021},
      eprint={2106.09685},
      archivePrefix={arXiv},
      primaryClass={cs.CL},
      url={https://arxiv.org/abs/2106.09685},
}

@misc{yang2025xhumanoidrobotizehumanvideos,
      title={X-Humanoid: Robotize Human Videos to Generate Humanoid Videos at Scale}, 
      author={Pei Yang and Hai Ci and Yiren Song and Mike Zheng Shou},
      year={2025},
      eprint={2512.04537},
      archivePrefix={arXiv},
      primaryClass={cs.CV},
      url={https://arxiv.org/abs/2512.04537}, 
}

@article{song2025mitty,
  title={Mitty: Diffusion-based Human-to-Robot Video Generation},
  author={Song, Yiren and Liu, Cheng and Mao, Weijia and Shou, Mike Zheng},
  journal={arXiv preprint arXiv:2512.17253},
  year={2025}
}

@inproceedings{lepert2026masquerade,
  title={{Masquerade}: Learning from In-the-wild Human Videos using Data-Editing},
  author={Lepert, Marion and Fang, Jiaying and Bohg, Jeannette},
  booktitle={2026 IEEE International Conference on Robotics and Automation (ICRA)},
  year={2026},
  organization={IEEE},
  url={https://arxiv.org/abs/2508.09976}
}

@article{ku2024anyv2v,
  title={Anyv2v: A tuning-free framework for any video-to-video editing tasks},
  author={Ku, Max and Wei, Cong and Ren, Weiming and Yang, Harry and Chen, Wenhu},
  journal={arXiv preprint arXiv:2403.14468},
  year={2024},
  eprint={2403.14468},
  archivePrefix={arXiv},
  primaryClass={cs.CV},
  url={https://arxiv.org/abs/2403.14468}
}

@article{ci2025h2r,
  title={H2R-Grounder: A Paired-Data-Free Paradigm for Translating Human Interaction Videos into Physically Grounded Robot Videos},
  author={Ci, Hai and Liu, Xiaokang and Yang, Pei and Song, Yiren and Shou, Mike Zheng},
  journal={arXiv preprint arXiv:2512.09406},
  year={2025}
}

@inproceedings{todorov2012mujoco,
  title={MuJoCo: A physics engine for model-based control},
  author={Todorov, Emanuel and Erez, Tom and Tassa, Yuval},
  booktitle={2012 IEEE/RSJ International Conference on Intelligent Robots and Systems},
  pages={5026--5033},
  year={2012},
  organization={IEEE},
  doi={10.1109/IROS.2012.6386109}
}

@article{ravi2024sam2,
  title={SAM 2: Segment Anything in Images and Videos},
  author={Ravi, Nikhila and Gabeur, Valentin and Hu, Yuan-Ting and others},
  journal={arXiv preprint arXiv:2408.00714},
  url={https://arxiv.org/abs/2408.00714},
  year={2024}
}

@misc{xie2025human2robotlearningrobotactions,
      title={Human2Robot: Learning Robot Actions from Paired Human-Robot Videos}, 
      author={Sicheng Xie and Haidong Cao and Zejia Weng and Zhen Xing and Haoran Chen and Shiwei Shen and Jiaqi Leng and Zuxuan Wu and Yu-Gang Jiang},
      year={2025},
      eprint={2502.16587},
      archivePrefix={arXiv},
      primaryClass={cs.RO},
      url={https://arxiv.org/abs/2502.16587}, 
}

@misc{yuan2025roboengineplugandplayrobotdata,
      title={RoboEngine: Plug-and-Play Robot Data Augmentation with Semantic Robot Segmentation and Background Generation}, 
      author={Chengbo Yuan and Suraj Joshi and Shaoting Zhu and Hang Su and Hang Zhao and Yang Gao},
      year={2025},
      eprint={2503.18738},
      archivePrefix={arXiv},
      primaryClass={cs.RO},
      url={https://arxiv.org/abs/2503.18738}, 
}

@misc{wu2025qwenimagetechnicalreport,
      title={Qwen-Image Technical Report},
      author={Chenfei Wu and Jiahao Li and Jingren Zhou and others},
      year={2025},
      eprint={2508.02324},
      archivePrefix={arXiv},
      primaryClass={cs.CV},
      url={https://arxiv.org/abs/2508.02324},
}

@ARTICLE{1284395,
  author={Zhou Wang and Bovik, A.C. and Sheikh, H.R. and Simoncelli, E.P.},
  journal={IEEE Transactions on Image Processing}, 
  title={Image quality assessment: from error visibility to structural similarity}, 
  year={2004},
  volume={13},
  number={4},
  pages={600-612},
  doi={10.1109/TIP.2003.819861}
}

@misc{zhang2018unreasonableeffectivenessdeepfeatures,
      title={The Unreasonable Effectiveness of Deep Features as a Perceptual Metric}, 
      author={Richard Zhang and Phillip Isola and Alexei A. Efros and Eli Shechtman and Oliver Wang},
      year={2018},
      eprint={1801.03924},
      archivePrefix={arXiv},
      primaryClass={cs.CV},
      url={https://arxiv.org/abs/1801.03924}, 
}

@misc{chen2026vinounifiedvisualgenerator,
      title={VINO: A Unified Visual Generator with Interleaved OmniModal Context}, 
      author={Junyi Chen and Tong He and Zhoujie Fu and Pengfei Wan and Kun Gai and Weicai Ye},
      year={2026},
      eprint={2601.02358},
      archivePrefix={arXiv},
      primaryClass={cs.CV},
      url={https://arxiv.org/abs/2601.02358}, 
}

@misc{kiwiedit,
      title={Kiwi-Edit: Versatile Video Editing via Instruction and Reference Guidance}, 
      author={Yiqi Lin and Guoqiang Liang and Ziyun Zeng and Zechen Bai and Yanzhe Chen and Mike Zheng Shou},
      year={2026},
      eprint={2603.02175},
      archivePrefix={arXiv},
      primaryClass={cs.CV},
      url={https://arxiv.org/abs/2603.02175}, 
}

@misc{pan2026omniweavingunifiedvideogeneration,
      title={OmniWeaving: Towards Unified Video Generation with Free-form Composition and Reasoning}, 
      author={Kaihang Pan and Qi Tian and Jianwei Zhang and others},
      year={2026},
      eprint={2603.24458},
      archivePrefix={arXiv},
      primaryClass={cs.CV},
      url={https://arxiv.org/abs/2603.24458}, 
}

@misc{litman2026editctrldisentangledlocalglobal,
      title={EditCtrl: Disentangled Local and Global Control for Real-Time Generative Video Editing}, 
      author={Yehonathan Litman and Shikun Liu and Dario Seyb and Nicholas Milef and Yang Zhou and Carl Marshall and Shubham Tulsiani and Caleb Leak},
      year={2026},
      eprint={2602.15031},
      archivePrefix={arXiv},
      primaryClass={cs.CV},
      url={https://arxiv.org/abs/2602.15031}, 
}

@article{zhang2025region,
  title={Region-Constraint In-Context Generation for Instructional Video Editing},
  author={Zhang, Zhongwei and Long, Fuchen and Li, Wei and Qiu, Zhaofan and Liu, Wu and Yao, Ting and Mei, Tao},
  journal={https://arxiv.org/abs/2512.17650},
  year={2025}
}

@inproceedings{grauman2022ego4d,
  title     = {{Ego4D}: Around the World in 3,000 Hours of Egocentric Video},
  author    = {Grauman, Kristen and Westbury, Andrew and Byrne, Eugene and others},
  booktitle = {IEEE/CVF Conference on Computer Vision and Pattern Recognition (CVPR)},
  year      = {2022}
}

@inproceedings{goyal2017somethingsomething,
  title     = {The ``Something Something'' Video Database for Learning and Evaluating Visual Common Sense},
  author    = {Goyal, Raghav and Kahou, Samira Ebrahimi and Michalski, Vincent and Materzy{\'n}ska, Joanna and Westphal, Susanne and Kim, Heuna and Haenel, Valentin and Fr{\"u}nd, Ingo and Yianilos, Peter and Mueller-Freitag, Moritz and others},
  booktitle = {IEEE International Conference on Computer Vision (ICCV)},
  year      = {2017}
}

@inproceedings{nair2022r3m,
  title     = {{R3M}: A Universal Visual Representation for Robot Manipulation},
  author    = {Nair, Suraj and Rajeswaran, Aravind and Kumar, Vikash and Finn, Chelsea and Gupta, Abhinav},
  booktitle = {Conference on Robot Learning (CoRL)},
  year      = {2022}
}

@inproceedings{ma2023vip,
  title     = {{VIP}: Towards Universal Visual Reward and Representation via Value-Implicit Pre-Training},
  author    = {Ma, Yecheng Jason and Sodhani, Shagun and Jayaraman, Dinesh and Bastani, Osbert and Kumar, Vikash and Zhang, Amy},
  booktitle = {International Conference on Learning Representations (ICLR)},
  year      = {2023}
}

@inproceedings{zakka2021xirl,
  title     = {{XIRL}: Cross-embodiment Inverse Reinforcement Learning},
  author    = {Zakka, Kevin and Zeng, Andy and Florence, Pete and Tompson, Jonathan and Bohg, Jeannette and Dwibedi, Debidatta},
  booktitle = {Conference on Robot Learning (CoRL)},
  year      = {2021}
}

@inproceedings{bahl2023affordances,
  title     = {Affordances from Human Videos as a Versatile Representation for Robotics},
  author    = {Bahl, Shikhar and Mendonca, Russell and Chen, Lili and Jain, Unnat and Pathak, Deepak},
  booktitle = {IEEE/CVF Conference on Computer Vision and Pattern Recognition (CVPR)},
  year      = {2023}
}

@inproceedings{shan2020understanding,
  title     = {Understanding Human Hands in Contact at Internet Scale},
  author    = {Shan, Dandan and Geng, Jiaqi and Shu, Michelle and Fouhey, David F.},
  booktitle = {IEEE/CVF Conference on Computer Vision and Pattern Recognition (CVPR)},
  year      = {2020}
}

@inproceedings{wang2023mimicplay,
  title     = {{MimicPlay}: Long-Horizon Imitation Learning by Watching Human Play},
  author    = {Wang, Chen and Fan, Linxi and Sun, Jiankai and Zhang, Ruohan and Fei-Fei, Li and Xu, Danfei and Zhu, Yuke and Anandkumar, Anima},
  booktitle = {Conference on Robot Learning (CoRL)},
  year      = {2023}
}

@inproceedings{he2015deepresiduallearningimage,
  title     = {Deep Residual Learning for Image Recognition},
  author    = {He, Kaiming and Zhang, Xiangyu and Ren, Shaoqing and Sun, Jian},
  booktitle = {IEEE Conference on Computer Vision and Pattern Recognition (CVPR)},
  pages     = {770--778},
  year      = {2016}
}

@article{MANO:SIGGRAPHASIA:2017,
      title = {Embodied Hands: Modeling and Capturing Hands and Bodies Together},
      author = {Romero, Javier and Tzionas, Dimitrios and Black, Michael J.},
      journal = {ACM Transactions on Graphics, (Proc. SIGGRAPH Asia)},
      volume = {36},
      number = {6},
      series = {245:1--245:17},
      month = nov,
      year = {2017},
      month_numeric = {11}
  }

@inproceedings{10.1007/978-3-030-58452-8_28,
      author = {Terzakis, George and Lourakis, Manolis},
      title = {A Consistently Fast and Globally Optimal Solution to the Perspective-n-Point Problem},
      year = {2020},
      isbn = {978-3-030-58451-1},
      publisher = {Springer-Verlag},
      address = {Berlin, Heidelberg},
      url = {https://doi.org/10.1007/978-3-030-58452-8_28},
      doi = {10.1007/978-3-030-58452-8_28},
      booktitle = {Computer Vision – ECCV 2020: 16th European Conference, Glasgow, UK, August 23–28, 2020, Proceedings, Part I},
      pages = {478–494},
      numpages = {17},
      location = {Glasgow, United Kingdom}
}

@misc{iqbal2018handposeestimationlatent,
      title={Hand Pose Estimation via Latent 2.5D Heatmap Regression},
      author={Umar Iqbal and Pavlo Molchanov and Thomas Breuel and Juergen Gall and Jan Kautz},
      year={2018},
      eprint={1804.09534},
      archivePrefix={arXiv},
      primaryClass={cs.CV},
      url={https://arxiv.org/abs/1804.09534},
}

@misc{schulman2017proximalpolicyoptimizationalgorithms,
      title={Proximal Policy Optimization Algorithms}, 
      author={John Schulman and Filip Wolski and Prafulla Dhariwal and Alec Radford and Oleg Klimov},
      year={2017},
      eprint={1707.06347},
      archivePrefix={arXiv},
      primaryClass={cs.LG},
      url={https://arxiv.org/abs/1707.06347}, 
}

@misc{ma2023livlanguageimagerepresentationsrewards,
      title={LIV: Language-Image Representations and Rewards for Robotic Control}, 
      author={Yecheng Jason Ma and William Liang and Vaidehi Som and Vikash Kumar and Amy Zhang and Osbert Bastani and Dinesh Jayaraman},
      year={2023},
      eprint={2306.00958},
      archivePrefix={arXiv},
      primaryClass={cs.RO},
      url={https://arxiv.org/abs/2306.00958}, 
}

@inproceedings{handdgp2024,
  title={{HandDGP}: Camera-Space Hand Mesh Prediction with Differentiable Global Positioning},
  author={Valassakis, Eugene and Garcia-Hernando, Guillermo},
  booktitle={Proceedings of the European Conference on Computer Vision (ECCV)},
  year={2024},
}

@misc{chen2026robodojo,
  title = {{RoboDojo}: A Unified Sim-and-Real Benchmark for Comprehensive Evaluation of Generalist Robot Manipulation Policies},
  author = {Chen, Tianxing and Chen, Yue and Li, Zixuan and others},
  year = {2026},
  eprint = {2607.04434},
  archivePrefix = {arXiv},
  primaryClass = {cs.RO},
  url = {https://arxiv.org/abs/2607.04434},
}

@misc{ren2026camerahand,
  title={Estimating Accurate Hand Pose in Camera Space with Vision Transformer},
  author={Kaiwen Ren and Yiran Jiang and Yongjing Ye and Shihong Xia},
  year={2026},
  eprint={2609.24424},
  archivePrefix={arXiv},
  primaryClass={cs.CV},
  doi={10.1145/3767308.3836423},
  url={https://arxiv.org/abs/2609.24424}
}
\endgroup
\clearpage
\appendix
\showrevisionsfalse
\setcounter{figure}{0}
\setcounter{table}{0}
\renewcommand{\thefigure}{S\arabic{figure}}
\renewcommand{\thetable}{S\arabic{table}}
\renewcommand{\theHfigure}{supp.\arabic{figure}}
\renewcommand{\theHtable}{supp.\arabic{table}}
\section*{\revision{Appendix}}
\section{RoboEdit-ADC Details}

\subsection{Camera Alignment and Depth Regularization}
\label{sec:supp-camera-depth}
\paragraph{Camera Alignment.}
VGGT predicts camera-from-world extrinsics
\(E_t=[R^{\mathrm{cw}}_t\mid t^{\mathrm{cw}}_t]\), camera intrinsics \(K_t\),
and depth maps \(D_t\). We convert each extrinsic into camera center
\(c_t=-{R^{\mathrm{cw}}_t}^{\top}t^{\mathrm{cw}}_t\) and orientation
\(R^{\mathrm{wc}}_t={R^{\mathrm{cw}}_t}^{\top}\).
Because the VGGT
reconstruction has an arbitrary scale and world frame, we align it to the
hand-object rendering frame with a global similarity transform
\((s,R_a,t_a)\).
We estimate this transform using standard closed-form SVD
alignment of sparse corresponding camera centers between the VGGT and rendering
coordinate systems.
The aligned camera and depth are
\(c'_t=sR_a c_t+t_a\), \(R'_t=R_aR^{\mathrm{wc}}_t\), and
\(\widetilde{D}_t=sD_t\), yielding \(C_t=(K_t,c'_t,R'_t)\).

\paragraph{Depth Regularization.}
For depth regularization, we use the aligned
depth map to correct the scale and depth of the monocular HaMeR hand
reconstruction before retargeting.
We select stable wrist and finger-base landmarks
from MANO~\citep{MANO:SIGGRAPHASIA:2017} as palm anchors and denote their set by
\(\mathcal{A}\). For anchor \(j\), we project its HaMeR camera-space position
\(H_{t,j}\) with \(K_t\), map the resulting pixel
\(\mathbf{u}_{t,j}=(u_{t,j},v_{t,j})\) to the resolution of
\(\widetilde D_t\), and take
the median positive, finite depth \(d_{t,j}\) within a \(5\times5\)
neighborhood. We then back-project this observation into a metric 3D anchor:
\[
p^{D}_{t,j}
=d_{t,j}K_t^{-1}
\bigl[u_{t,j},v_{t,j},1\bigr]^{\top}.
\]
Invalid samples are excluded. From the valid set \(\mathcal{A}_t\), we compute
\(\bar z_t^{D}=\mathrm{median}_{j\in\mathcal{A}_t}
(p^{D}_{t,j})_z\) and
\(\bar z_t^{H}=\mathrm{median}_{j\in\mathcal{A}_t}(H_{t,j})_z\), giving
\(\alpha_t=\bar z_t^{D}/\bar z_t^{H}\). We rescale the complete camera-space
hand about the camera center as \(\widetilde H_t=\alpha_t H_t\), applying the
same factor to its global translation while leaving the MANO pose, shape, and
relative articulation unchanged.
This camera-centered isotropic scaling aligns the metric
palm depth while preserving the original image projection.

\subsection{Physics-guided Refinement}
\label{sec:supp-retargeting}

\begingroup

\paragraph{Refinement objective.}
We adjust wrist poses and finger joints around the kinematic retargeting
result, keeping object motion fixed. Let \(t\) index frames and \(h\)
index hands; all sums below run over valid hand frames. The tracking,
penetration, contact, and temporal terms have weights 0.02, 1, 20, and 2,
respectively, in the main paper's refinement objective.

\paragraph{Tracking and temporal losses.}
Let \(u_{t,h}\) be the wrist and finger correction vector, with each
component divided by its allowed correction bound (given below).
Let \(\bar u_{t,h}\) be the mean normalized correction of the valid
adjacent frames. We use
\begin{align*}
\mathcal L_{\mathrm{track}}&=\sum_{t,h}\|u_{t,h}\|_2^2,\\
\mathcal L_{\mathrm{temp}}&=\sum_{t,h}\|u_{t,h}-\bar u_{t,h}\|_2^2.
\end{align*}
Here, \(\|\cdot\|_2\) is the Euclidean norm. Tracking limits changes from
the kinematic reference; temporal regularization keeps these changes
consistent across neighboring frames. A frame without valid neighbors
has no temporal penalty.

\paragraph{Penetration loss.}
For each robot-link convex hull, we select the closest object convex part
if their signed distance is below 2\,mm. These pairs form
\(\mathcal P_{t,h}\). For each pair \(c\) in the current hand frame,
let \(d_c\) be its signed distance (negative for penetration) and \(p_c\)
its squared clearance penalty:
\begin{align*}
p_c&=\left(\frac{[1\,\mathrm{mm}-d_c]_+}{s_d}\right)^2,\\
\mathcal L_{\mathrm{geo}}&=\sum_{t,h}\left(
\underset{c\in\mathcal P_{t,h}}{\operatorname{mean}}p_c
+2\max_{c\in\mathcal P_{t,h}}p_c\right).
\end{align*}
Here, \([x]_+=\max(x,0)\), and \(s_d=10\,\mathrm{mm}\) normalizes distances.
This penalizes overlap with a 1\,mm clearance margin and gives extra weight
to the worst pair. A hand frame with no selected pairs contributes zero.

\paragraph{Contact preservation.}
We set \(I_{t,h}=1\) when the nearest source-human fingertip is within
20\,mm of the object surface, and zero otherwise. Let \(g_{t,h}\) be the
nearest robot-fingertip distance to that surface. We use
\[
\mathcal L_{\mathrm{contact}}
=\sum_{t,h}I_{t,h}\left(\frac{[g_{t,h}-8\,\mathrm{mm}]_+}{s_d}\right)^2.
\]
Thus, robot distances above 8\,mm are penalized only on source-contact frames.

\paragraph{Optimization.}
We use one forward and one backward sweep, with at most 12 bounded
Jacobian-based least-squares updates per hand frame and backtracking.
Wrist corrections are limited to 30\,mm and 0.14\,rad; finger corrections
to 0.6\,rad for rotational joints or 20\,mm for translational joints,
subject to joint limits and gripper couplings. Updates preserve existing
robot contact on source-contact frames. We apply Gaussian smoothing to
corrections within valid segments, with a standard deviation of 3 frames,
and blend with the unsmoothed corrections under penetration and contact
constraints. We accept the refined trajectory only if its mean penetration,
95th-percentile penetration, and missed-contact count (Section~\ref{sec:supp-adc-qc})
do not increase relative to the kinematic reference.

\subsection{Trajectory Quality Control}
\label{sec:supp-adc-qc}

\paragraph{Quality Control Standards.}
We measure penetration as the maximum robot--object overlap depth in
each frame and use its 95th percentile over the clip as the penetration
score. For clips containing source contact, the contact-miss rate is
\[
r_{\mathrm{miss}}=
\frac{\sum_{t,h}I_{t,h}\,\mathbf{1}[g_{t,h}>20\,\mathrm{mm}]}
     {\sum_{t,h}I_{t,h}}.
\]
Here, \(t\), \(h\), \(I_{t,h}\), and \(g_{t,h}\) follow
Section~\ref{sec:supp-retargeting}; \(\mathbf{1}[\cdot]\) is 1 when its
condition holds and 0 otherwise. The denominator counts source-contact
hand frames in the clip; the numerator counts those for which the robot
fingertip distance exceeds 20\,mm. We retain clips with a penetration score
of at most 20\,mm and \(r_{\mathrm{miss}}\leq20\%\).

\paragraph{Full-data evaluation.}
We apply refinement and quality control to the 145,459
real-scene clips. Refinement reduces mean penetration from 12.20 to 2.69\,mm and
the contact-miss rate from 1.439\% to 0.103\% on the same assessed clips.
Quality control retains 139,058 clips (95.60\% of the clips inputs)
and rejects 6,401 assessed clips. The retained subset has mean penetration
2.33\,mm and contact-miss rate 0.047\%.
To separate refinement from filtering, the same retained subset has
mean penetration 11.66\,mm and contact-miss rate 1.353\% before refinement.
Table~\ref{tab:supp-adc-quality-breakdown} reports results by source and
embodiment.
\endgroup

\begin{table*}[!t]
\centering
\caption{RoboEdit-ADC refinement and quality control by source and target embodiment. B: before refinement; R: after refinement, before filtering; Q: after quality control. Input and retained counts refer to real-scene clips.}
\label{tab:supp-adc-quality-breakdown}
\begingroup
\small
\setlength{\tabcolsep}{3pt}
\begin{tabular*}{\textwidth}{@{\extracolsep{\fill}}lrrrcccccc@{}}
\toprule
& \multicolumn{3}{c}{Clips and retention} & \multicolumn{3}{c}{Mean penetration (mm) $\downarrow$} & \multicolumn{3}{c}{Contact miss (\%) $\downarrow$} \\
\cmidrule(lr){2-4}\cmidrule(lr){5-7}\cmidrule(l){8-10}
Group & Input & Retained & (\%) & B & R & Q & B & R & Q \\
\midrule
\textbf{All real-scene clips} & 145,459 & 139,058 & 95.60 & 12.20 & 2.69 & 2.33 & 1.439 & 0.103 & 0.047 \\
\midrule
\multicolumn{10}{l}{\textit{By source}} \\
DexYCB & 56,000 & 54,746 & 97.76 & 10.31 & 1.77 & 1.62 & 1.402 & 0.048 & 0.029 \\
HOT3D & 29,659 & 28,661 & 96.64 & 11.12 & 2.18 & 1.95 & 1.629 & 0.206 & 0.092 \\
H2O & 6,630 & 6,556 & 98.88 & 8.84 & 1.14 & 1.09 & 1.451 & 0.028 & 0.014 \\
GigaHands & 9,760 & 8,561 & 87.72 & 18.59 & 4.33 & 2.91 & 2.299 & 0.242 & 0.075 \\
TACO & 43,410 & 40,534 & 93.37 & 14.46 & 4.07 & 3.63 & 1.153 & 0.059 & 0.033 \\
\midrule
\multicolumn{10}{l}{\textit{By embodiment}} \\
Inspire & 24,197 & 23,582 & 97.46 & 9.85 & 1.77 & 1.57 & 1.342 & 0.062 & 0.041 \\
XHand & 24,197 & 23,304 & 96.31 & 13.13 & 2.75 & 2.50 & 1.342 & 0.074 & 0.034 \\
Ability & 24,197 & 23,502 & 97.13 & 11.64 & 2.34 & 2.16 & 1.191 & 0.099 & 0.048 \\
SCHUNK SVH & 24,197 & 23,801 & 98.36 & 11.33 & 2.43 & 2.32 & 1.398 & 0.069 & 0.044 \\
Allegro & 24,197 & 21,212 & 87.66 & 15.86 & 4.73 & 3.74 & 1.873 & 0.165 & 0.087 \\
Unitree Dex3 & 12,237 & 11,928 & 97.47 & 11.65 & 2.00 & 1.82 & 1.153 & 0.076 & 0.022 \\
Panda gripper & 12,237 & 11,729 & 95.85 & 11.17 & 2.22 & 1.81 & 1.868 & 0.273 & 0.032 \\
\bottomrule
\end{tabular*}
\endgroup
\end{table*}

\section{3D Robot-State Decoder Details}
\label{sec:supp-decoder}

\paragraph{Decoder Architecture.}
A shared ResNet-34~\citep{he2015deepresiduallearningimage} feature pyramid
extracts local and global features.
For each robot hand, a palm-anchor head predicts heatmaps and
visibility for eight predefined 2D palm anchors, a spatial fingertip heatmap head localizes up to five
fingertips, and an MLP state head predicts palm-frame 3D fingertip coordinates
and normalized joint configurations.
A shared camera head predicts intrinsics.
SQPnP~\citep{10.1007/978-3-030-58452-8_28} recovers the camera-space
wrist pose from the predicted 2D palm anchors and predefined 3D palm geometry.
A temporal Transformer refines the framewise states over all 81 frames, and
embodiment-specific forward kinematics produces the complete camera-space
robot-hand trajectory.

\paragraph{Loss Functions.}
The decoder is trained with three objectives:
\[
\begin{array}{l}
\mathcal{L}_{\mathrm{anchor}}
=\lambda^{a}_{\mathrm{hm}}\mathcal{L}^{a}_{\mathrm{hm}}
+\lambda^{a}_{\mathrm{uv}}\mathcal{L}^{a}_{\mathrm{uv}}
+\lambda^{a}_{\mathrm{vis}}\mathcal{L}^{a}_{\mathrm{vis}},\\
\mathcal{L}_{\mathrm{spa}}
=\lambda_{\mathrm{2D}}\mathcal{L}_{\mathrm{2D}}
+\lambda^{s}_{\mathrm{palm}}\mathcal{L}_{\mathrm{palm}}
+\lambda_{\mathrm{cam}}\mathcal{L}_{\mathrm{cam}}\\
\hspace{15mm}
+\lambda^{s}_{\mathrm{qpos}}\mathcal{L}_{\mathrm{qpos}}
+\lambda_{K}\mathcal{L}_{K},\\
\mathcal{L}_{\mathrm{tmp}}
=\lambda^{t}_{\mathrm{palm}}\mathcal{L}_{\mathrm{palm}}
+\lambda^{t}_{\mathrm{qpos}}\mathcal{L}_{\mathrm{qpos}}
+\lambda_{\Delta\mathrm{palm}}\mathcal{L}_{\Delta\mathrm{palm}}\\
\hspace{15mm}
+\lambda_{\Delta\mathrm{qpos}}\mathcal{L}_{\Delta\mathrm{qpos}}
+\lambda_{\mathrm{res}}\mathcal{L}_{\mathrm{res}} .
\end{array}
\]
Here, each \(\lambda\) is a scalar weight for its supervision term;
\(a\), \(s\), and \(t\) distinguish anchor, spatial, and
temporal supervision, respectively.
The individual terms are defined as follows:

\noindent\textit{Anchor Heatmap Loss:}
\(\mathcal{L}^{a}_{\mathrm{hm}}\) uses a heatmap-distribution loss to supervise
the rigid palm-anchor heatmaps.

\noindent\textit{Anchor Coordinate Loss:}
\(\mathcal{L}^{a}_{\mathrm{uv}}\) uses Smooth-L1 loss on the 2D image
coordinates of the palm anchors.

\noindent\textit{Anchor Visibility Loss:}
\(\mathcal{L}^{a}_{\mathrm{vis}}\) uses binary cross-entropy to supervise
anchor visibility.

\noindent\textit{2D Fingertip Loss:}
\(\mathcal{L}_{\mathrm{2D}}\) combines a heatmap-distribution loss with
Smooth-L1 loss on the 2D fingertip coordinates.

\noindent\textit{3D Geometry Losses:}
\(\mathcal{L}_{\mathrm{palm}}\) and \(\mathcal{L}_{\mathrm{cam}}\) use
Smooth-L1 losses on 3D fingertip coordinates in the palm and camera frames,
respectively.

\noindent\textit{Joint-State Loss:}
\(\mathcal{L}_{\mathrm{qpos}}\) uses Smooth-L1 loss on normalized robot joint
states.

\noindent\textit{Camera-Intrinsics Loss:}
\(\mathcal{L}_{K}\) uses Smooth-L1 loss on the predicted camera intrinsics.

\noindent\textit{Temporal Motion Losses:}
\(\mathcal{L}_{\Delta\mathrm{palm}}\) and
\(\mathcal{L}_{\Delta\mathrm{qpos}}\) use Smooth-L1 losses to match
adjacent-frame fingertip and joint-state motion.

\noindent\textit{Temporal Residual Loss:}
\(\mathcal{L}_{\mathrm{res}}\) applies an L2 penalty to the temporal
corrections.

Unavailable fingertips, joints, and frames are excluded using validity masks.
\(\mathcal{L}_{\mathrm{anchor}}\), \(\mathcal{L}_{\mathrm{spa}}\), and
\(\mathcal{L}_{\mathrm{tmp}}\) are the weighted combinations shown above.

\section{RoboEdit-14M Statistics}
\label{sec:supp-dataset}

\begin{figure*}[!t]
\centering
\includegraphics[width=5.5in]{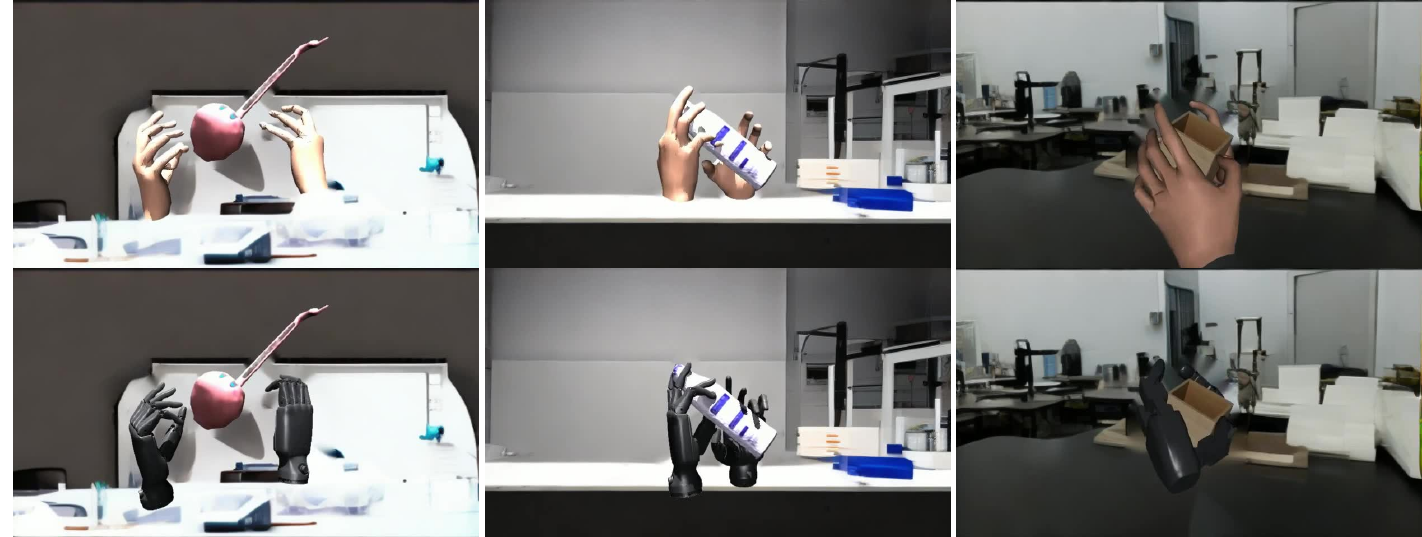}
\caption{Synthetic paired-video examples. Each column shows
an aligned synthetic human frame and its robot target under the
same scene, object state, and camera view.}
\label{fig:supp-synthetic-augmentation}

\end{figure*}

Table~\ref{tab:supp-adc-quality-breakdown} summarizes the effects
of refinement and quality control. Tables~\ref{tab:supp-source-counts} and
\ref{tab:supp-embodiment-counts} summarize RoboEdit-14M by source dataset
and target embodiment, respectively.
Fig.~\ref{fig:supp-synthetic-augmentation} shows representative synthetic
human/robot pairs.

\begin{table*}[!t]
\begin{minipage}[t]{0.48\linewidth}
\vspace{0pt}
\centering
\caption{{RoboEdit-14M source distribution, with duration computed at 30 FPS.}
}
\label{tab:supp-source-counts}
\small
\setlength{\tabcolsep}{2.5pt}
\begin{tabular*}{\linewidth}{@{\extracolsep{\fill}}lrrr@{}}
\toprule
Source dataset & Paired clips & Paired frames & Duration (h) \\
\midrule
DexYCB & 67,200 & 5,443,200 & 50.40 \\
GigaHands & 11,715 & 948,915 & 8.79 \\
H2O & 7,955 & 644,355 & 5.97 \\
HOT3D & 35,591 & 2,882,871 & 26.69 \\
TACO & 52,086 & 4,218,966 & 39.06 \\
\midrule
Total & 174,547 & 14,138,307 & 130.91 \\
\bottomrule
\end{tabular*}
\small

\end{minipage}\hfill
\begin{minipage}[t]{0.48\linewidth}
\vspace{0pt}
\centering
\caption{{RoboEdit-14M distribution by target embodiment.}}
\label{tab:supp-embodiment-counts}
\small
\setlength{\tabcolsep}{2.5pt}
\begin{tabular*}{\linewidth}{@{\extracolsep{\fill}}lrrr@{}}
\toprule
Embodiment & Real & Synthetic & Total \\
\midrule
Ability & 23,502 & 5,534 & 29,036 \\
Allegro & 21,212 & 7,824 & 29,036 \\
Unitree Dex3 & 11,928 & 2,756 & 14,684 \\
Inspire & 23,582 & 5,454 & 29,036 \\
Panda gripper & 11,729 & 2,955 & 14,684 \\
SCHUNK SVH & 23,801 & 5,234 & 29,035 \\
XHand & 23,304 & 5,732 & 29,036 \\
\midrule
Total & 139,058 & 35,489 & 174,547 \\
\bottomrule
\end{tabular*}
\small

\end{minipage}
\end{table*}

\begin{figure*}[!t]
\centering
\includegraphics[width=5.5in]{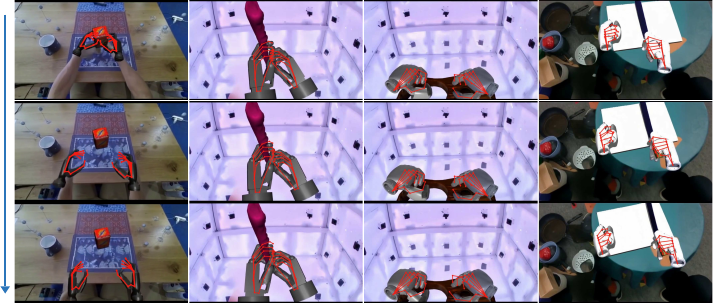}
\caption{Additional 3D Robot-State Decoder results on
RoboEdit-Trans edited videos. Red overlays denote predicted
camera-space hand states.
}
\label{fig:supp-decoder}
\end{figure*}

\section{Training Details}
\label{sec:supp-training}

\paragraph{Computing Infrastructure.}
Experiments were conducted on Ubuntu 22.04.5 workstations
equipped with two AMD EPYC 9354 CPUs, 1.5\,TiB of host memory, and eight
NVIDIA H100 NVL GPUs, each with 94\,GiB of memory.

\paragraph{Backbone and Cross-Embodiment Adaptation.}
We first train the video-editing backbone for approximately 10K
steps, updating 283.4M parameters with AdamW at a learning rate of
\(5\times10^{-5}\) and an effective batch size of 12 across eight GPUs. We then
freeze the backbone and jointly train LoRA adapters on selected attention and
feed-forward projections together with residual bottleneck adapters in the
final ten transformer blocks, using an effective batch size of 16 across four
GPUs. The LoRA and residual-adapter branches contain 2.5M and 7.9M trainable
parameters, respectively.

\paragraph{Baseline Backbone Configuration.}
\label{sec:supp-backbone-config}

The VACE baselines use pretrained \texttt{Wan2.1-VACE-1.3B}, with UMT5-XXL
and the Wan2.1 VAE.
Inference uses 81 frames at 480p with 20-step UniPC sampling, guidance scale
5.0, and flow shift 16.
RoboEdit-Trans initializes its diffusion transformer and image encoder from
\texttt{Wan2.1-Fun-1.3B-InP} and its text encoder and VAE from
\texttt{Wan2.1-T2V-1.3B} before the task adaptation described above.
\par
\paragraph{Masquerade Setting.}
Masquerade~\citep{lepert2026masquerade} natively supports the Robotiq 2F-85 parallel-jaw gripper. We extend it to our seven robot hands using their robot assets and adapting the pose-to-joint mapping. Of the 300 test cases, Masquerade produces 227 complete outputs, 57 partial or partly black outputs, and 16 failures.

\paragraph{3D Robot-State Decoder.}

We first train the palm detector and framewise spatial estimator
on rendered and composited robot frames. We then freeze both components and
train temporal refinement on complete 81-frame sequences.
We train the spatial and temporal stages for approximately
8.9K and 1.5K optimization steps, respectively; both stages use AdamW with a
learning rate of \(2\times10^{-5}\).

\paragraph{Qwen-Image-Edit.}
We fine-tune Qwen-Image-Edit for approximately 4.9K optimization steps on aligned
image pairs sampled from RoboEdit-14M. At
inference, it generates robot references at indices
\(\{0,10,\ldots,80\}\), after which RoboEdit-Trans produces an 81-frame video.

\paragraph{Randomness Control.}

For reproducibility, we fix the random seeds used for data
sampling, decoder training, and model inference. RoboEdit-Trans data sampling
initializes its NumPy generator with 20260531 plus the epoch index. Decoder
training initializes the Python, NumPy, and PyTorch generators with 20260705
plus the distributed rank. RoboEdit-Trans, Qwen-Image-Edit, and the original video-editing baselines use deterministic per-sample PyTorch seeds derived from 42, 20260629, and 20260624, respectively.

\begingroup

\section{{User Study}}
\label{app:user-study}
{We sample 30 cases from the 300 test cases. Each evaluation contains 60 comparison groups, and each comparison group pairs RoboEdit-Trans with one randomly selected, distinct baseline. Method names are hidden, and comparison order and A/B placement are randomized. The source video and target robot reference accompany each pair. Respondents choose A or B for overall quality, action/interaction preservation, robot appearance consistency, and contact plausibility. Table~\ref{tab:supp-user-study} reports preference for RoboEdit-Trans across 240 comparisons.}

We considered this voluntary video-rating task minimal-risk and did not seek ethics review. Participants consented and could stop at any time.

\begin{table}[htbp]
\centering
\caption{{Preference for RoboEdit-Trans (\%). Action, Robot, and Contact correspond to the three criteria following overall quality.}}
\label{tab:supp-user-study}
\begingroup
\small
\color{black}
\setlength{\tabcolsep}{3pt}
\begin{tabular*}{\columnwidth}{@{\extracolsep{\fill}}lrrrr@{}}
\toprule
Baseline & Overall & Action & Robot & Contact \\
\midrule
VACE & 80.0 & 82.2 & 82.2 & 82.2 \\
UniVideo & 94.3 & 94.3 & 97.1 & 94.3 \\
VINO & 100.0 & 100.0 & 100.0 & 97.7 \\
Kiwi-Edit & 82.4 & 82.4 & 100.0 & 82.4 \\
OmniWeaving & 85.0 & 85.0 & 95.0 & 80.0 \\
EditCtrl & 100.0 & 100.0 & 100.0 & 100.0 \\
AnyV2V & 100.0 & 100.0 & 100.0 & 100.0 \\
ReCo & 100.0 & 100.0 & 93.8 & 100.0 \\
Mitty & 100.0 & 100.0 & 100.0 & 100.0 \\
Masquerade & 100.0 & 100.0 & 100.0 & 100.0 \\
\midrule
Pooled & 92.9 & 93.3 & 95.4 & 92.5 \\
\bottomrule
\end{tabular*}
\endgroup
\end{table}

\color{black}
\begin{table*}[!t]
\centering
\color{black}
\captionsetup{labelfont={color=black},textfont={color=black}}

\caption{Runtime breakdown for RoboEdit-ADC and RoboEdit-Trans on one H100 GPU.
}
\label{tab:supp-runtime}
\small
\setlength{\tabcolsep}{4pt}
\renewcommand{\arraystretch}{1.00}
\begin{tabular*}{\linewidth}{@{\extracolsep{\fill}}llr@{}}
\toprule
Method & Stage & Time (s) \\
\midrule
RoboEdit-ADC & Object Segment & 48.17 \\
 & Object Pose & 59.25 \\
 & Object Mesh & 125.58 \\
 & Hand Pose & 251.92 \\
 & Camera State & 104.43 \\
 & Kinematic Retargeting & 111.50 \\
 & Depth Alignment & 18.66 \\
 & Physics Refinement & 313.97 \\
 & Inpainting & 129.08 \\
 & Foreground Rendering & 43.06 \\
 & \textbf{RoboEdit-ADC total} & \textbf{1205.63} \\
\midrule
Qwen-Image-Edit & Reference generation (9 frames) & 267.71 \\
\noalign{\vskip\aboverulesep
\hbox to\linewidth{\xleaders\hbox{\rule{3pt}{0.4pt}\kern2pt}\hfill}
\vskip\belowrulesep}
RoboEdit-Trans & Input preparation and encoding & 14.05 \\
 & Video denoising (20 steps) & 134.22 \\
 & VAE decoding & 7.92 \\
 & 3D Robot-State Decoder & 75.09 \\
 & \textbf{RoboEdit-Trans subtotal} & \textbf{231.28} \\
\midrule
Qwen-Image-Edit + RoboEdit-Trans & \textbf{Total} & \textbf{498.99} \\
\bottomrule
\end{tabular*}
\end{table*}

\section{Runtime Analysis}
\label{app:runtime}

Table~\ref{tab:supp-runtime} reports the stage-by-stage runtime of
RoboEdit-ADC curation,
Qwen reference keyframe generation, and RoboEdit-Trans inference
averaged over 300 benchmark cases on one NVIDIA H100 NVL GPU. Generating nine
Qwen references with 40 steps per image takes 267.71\,s. Together with
RoboEdit-Trans inference, reference generation, video editing, and 3D state
decoding take 498.99\,s (8.3\,min) in total, substantially more efficient than
the 20.0\,min required by RoboEdit-ADC.

\par
\endgroup

\section{Additional Qualitative Results}
\label{sec:supp-qualitative}

Figure~\ref{fig:supp-qualitative} extends the main-paper
comparison with six additional interaction examples. Figure~\ref{fig:supp-decoder} shows additional decoded robot states
from edited videos, and
Figure~\ref{fig:supp-rollouts} presents additional real-robot
results on YCB-object manipulation with both end effectors.

\section{In-the-Wild Evaluation}
\label{sec:supp-in-the-wild}
We collect 150 human manipulation clips, comprising 75 fixed-camera and 75 egocentric clips. RoboEdit-ADC constructs the robot counterparts; RoboEdit-Trans performs inference to generate the corresponding robot videos. Qualitative results are shown in Figure~\ref{fig:in-the-wild}.

\section{Trajectory-Conditioned Control and Real-Robot Deployment}
\label{app:trajectory-control}
\label{sec:supp-control}

\begingroup
\setlength{\parskip}{3pt}
\setlength{\abovedisplayskip}{4pt plus 2pt minus 2pt}
\setlength{\belowdisplayskip}{4pt plus 2pt minus 2pt}
\setlength{\abovedisplayshortskip}{0pt plus 2pt}
\setlength{\belowdisplayshortskip}{3pt plus 2pt minus 2pt}

We evaluate whether the robot-hand trajectories recovered by
the 3D Robot-State Decoder are physically executable and can supervise
downstream control.
{We compare direct tracking, residual control, and video-trained policies to assess how recovered trajectories support execution.}

\paragraph{Trajectory-conditioned residual control.}
For each interaction, the 3D Robot-State Decoder provides a
decoded robot-hand trajectory
\(\tau^h=\{h^\star_t\}_{t=1}^{T}\), which serves as the control reference;
\(h^\star_t\) denotes the reference palm or end-effector state. {Here, $t$ indexes control steps and $T$ is the trajectory length.}
{We transform references into world coordinates and interpolate at 30\,Hz. From robot/object states and reference look-ahead, the policy predicts a bounded residual to the nominal IK/servo command:}

{
\[
\begin{aligned}
a_t &= \pi_\phi(s_t,\tau^h),\\
\bar a_t &= 0.5\bar a_{t-1}+0.5a_t,\\
q^{\mathrm{cmd}}_t &= q^{\mathrm{nom}}_t+D\bar a_t.
\end{aligned}
\]
}

{Here, $\pi_\phi$ is the residual policy with parameters $\phi$; $a_t$ and $\bar a_t$ are its bounded, normalized action and the smoothed action, respectively. The state $s_t$ includes robot/object state and previous action. The nominal and corrected control commands are $q_t^{\mathrm{nom}}$ and $q_t^{\mathrm{cmd}}$. The nominal command includes velocity/gravity feedforward; the diagonal scaling matrix $D$ limits root corrections to $\pm4$\,cm and $\pm0.25$\,rad per axis, and fingers to actuator ranges.}

\paragraph{Control objective.}
{The reward combines reaching, contact, lifting, and holding with an action penalty:}

{
\[
r_t=r_t^{\mathrm{task}}-\frac{w_{\mathrm{act}}}{d}\|a_t\|_2^2.
\]
}

{Here, $r_t$ is the total reward, $r_t^{\mathrm{task}}$ is the task reward, $d$ is the action dimension, and $w_{\mathrm{act}}=0.01$ is the action-penalty weight. References condition commands and observations, not rewards.}

\begingroup
\begin{table}[!t]
\centering
\begingroup
\caption{Four YCB-object robot hand manipulation tasks success rates (\%) with residual control. Masq. denotes Masquerade-adapted.}
\label{tab:supp-sim-selected16}
\footnotesize
\setlength{\tabcolsep}{1.5pt}
\renewcommand{\arraystretch}{1.02}
\begin{tabular*}{\columnwidth}{@{\extracolsep{\fill}}llrrr@{}}
\toprule
Hand & Task & RoboEdit-ADC & RoboEdit-Trans & Masq. \\
\midrule
\multirow{5}{*}{Allegro} & Foam brick & 100.0 & 100.0 & 0.0 \\
 & Tomato can & 100.0 & 93.5 & 0.0 \\
 & Tuna can & 92.0 & 100.0 & 0.0 \\
 & Drill & 51.0 & 27.0 & 0.0 \\
 & \textbf{Mean} & \textbf{85.7} & \textbf{80.1} & \textbf{0.0} \\
\midrule
\multirow{5}{*}{XHand} & Foam brick & 100.0 & 100.0 & 0.0 \\
 & Tomato can & 95.0 & 100.0 & 0.0 \\
 & Tuna can & 0.0 & 100.0 & 0.0 \\
 & Pitcher & 44.5 & 19.0 & 0.0 \\
 & \textbf{Mean} & \textbf{59.9} & \textbf{79.8} & \textbf{0.0} \\
\midrule
\multirow{5}{*}{Schunk} & Tomato can & 100.0 & 83.0 & 0.0 \\
 & Tuna can & 82.0 & 96.5 & 0.0 \\
 & Drill & 100.0 & 26.0 & 0.0 \\
 & Pitcher & 21.0 & 17.5 & 0.0 \\
 & \textbf{Mean} & \textbf{75.8} & \textbf{55.8} & \textbf{0.0} \\
\midrule
\multirow{5}{*}{Panda gripper} & Tomato can & 97.0 & 97.5 & 0.0 \\
 & Sugar box & 79.0 & 95.0 & 0.0 \\
 & Foam brick & 100.0 & 32.0 & 0.0 \\
 & Scissors & 62.0 & 35.5 & 0.0 \\
 & \textbf{Mean} & \textbf{84.5} & \textbf{65.0} & \textbf{0.0} \\
 \midrule
 \textbf{Overall} & \textbf{Mean} & \textbf{76.5} & \textbf{70.2} & \textbf{0.0} \\
\bottomrule
\end{tabular*}
\endgroup
\end{table}
\endgroup

\paragraph{Simulation setup.}

{Table~\ref{tab:supp-sim-selected16} covers four grasp--lift--hold tasks per hand using seven YCB objects: foam brick, tomato soup can, tuna fish can, power drill, pitcher, sugar box, and scissors. Each task requires grasping the named object, lifting it, and maintaining the grasp.}

{In MuJoCo 4.0.1, we train one PPO policy per hand--task setting for 8M steps with 240 parallel environments. Training equally mixes RoboEdit-ADC kinematic, RoboEdit-Trans decoded, and Masquerade-adapted references without source labels. Each comparison uses the same policy weights, scene, servos, reward, and budget.}

{Training randomizes object translation by up to 5\,cm and yaw by $20^\circ$, with 20\% nominal states. Evaluation uses independent, matched initial states: 100 light trials (2\,cm, $10^\circ$) and 100 strong trials (5\,cm, $20^\circ$) per trajectory, transforming the reference with the object. After the demonstration, the reference adds a 10\,cm lift over 1\,s and a 2.5\,s hold. Success requires at least 5\,cm object elevation throughout the final 2\,s. We pool trials per task and equally average tasks per hand and all 16 settings overall.}

\paragraph{{Direct Tracking and Role of residual correction.}}
{To quantify the benefit of learned residual correction, we compare the same IK/servo tracker with and without residuals. Residual correction improves task success by 64.0 and 64.8 percentage points for RoboEdit-ADC and RoboEdit-Trans, respectively; Masquerade-adapted remains at 0.0\% because its native camera-scale mismatch prevents hand--object contact; residual corrections, limited to $\pm4$\,cm and $\pm0.25$\,rad per axis, cannot compensate for this discrepancy. Thus, the trajectories provide motion references, while residual feedback corrects grasp and contact errors (Figure~\ref{fig:supp-simulation-rollouts}).}

\paragraph{{Matched trajectory-source comparison.}}
{To assess downstream utility beyond video similarity, we compare reference sources under matched control (Table~\ref{tab:supp-sim-selected16}). RoboEdit-Trans approaches RoboEdit-ADC's success rate with lower processing time (about 8 versus 20 minutes), and far exceeds native Masquerade-adapted~\citep{lepert2026masquerade}. The latter uses its EPIC camera prior, whose metric pose misalignment limits object contact.}

\paragraph{{Learning policies directly from videos.}}
{We also train RGB-conditioned policies without online 3D references. We vary the video source while fixing RoboEdit-ADC action labels, and separately test RoboEdit-Trans videos with decoded labels and a no-video-pretraining control.}

{Under the current training budget of 1,500 behavior-cloning updates and six DAgger rounds (12 rollouts per case and 2,000 updates per round), success remains below $10\%$, with 189 evaluation episodes per configuration under the same lift-and-hold criterion.}

{Policies often contact the object but fail to grasp and hold it: palm-command errors and incomplete finger closure undermine contact, while teacher failures limit supervision. These results highlight the difficulty of learning reliable contact control from RGB observations in this setting and motivate explicit 3D motion references for residual control.}

\paragraph{Real-robot controller.}
\textcolor{blue}{}
We represent the trajectory relative to its first frame and
resample it to the control horizon. Given the current task state \(s_t\) and
the robot-hand tracking error \(h^\star_t-p_t\), the residual policy
\(\pi_\phi\) predicts a bounded correction to a nominal joint command computed
by inverse kinematics (IK):

\[
\begin{aligned}
a_t&=\pi_\phi\!\left(s_t,h^\star_t-p_t\right),\\
q^{\mathrm{cmd}}_t
&=\operatorname{clip}\!\left(
q^{\mathrm{IK}}_t+\rho_{\mathrm{act}}a_t,
q_{\min},q_{\max}
\right).
\end{aligned}
\]

Here, \(\pi_\phi\) is the residual policy with parameters
\(\phi\), \(s_t\) is the current robot and object task state, and \(p_t\) is
the current palm or end-effector state.
\(q^{\mathrm{IK}}_t\) is the nominal arm configuration computed by IK, and
\(q^{\mathrm{cmd}}_t\) is the resulting command sent to the low-level arm
controller;
\(a_t\in[-1,1]^7\) is the residual arm action, and
\(\rho_{\mathrm{act}}=0.02\,\mathrm{rad}\) is its scale.
\(q_{\min}\) and \(q_{\max}\) denote the Franka arm joint limits.
The operator \(\operatorname{clip}\) enforces these limits elementwise.
The low-level controller executes \(q^{\mathrm{cmd}}_t\),
whose outcome is evaluated through palm tracking and task success.

\paragraph{Control objective.}
The reward supervises tracking of the decoded
robot-hand trajectory while encouraging successful and smooth task execution:

\[
r_t=
w_{\mathrm{hand}}
\exp\!\left(-\beta\left\|p_t-h^\star_t\right\|_2\right)
+r_t^{\mathrm{task}}
-w_{\mathrm{act}}\left\|a_t\right\|_2^2 .
\]

Here, \(r_t\) is the total reward;
\(w_{\mathrm{hand}}\) and \(w_{\mathrm{act}}\) weight trajectory tracking and
action regularization, respectively; \(\beta\) controls sensitivity to the
Euclidean tracking error; and \(r_t^{\mathrm{task}}\) contains
rewards for grasp maintenance, object-motion consistency, and lift completion.
The decoder trajectory therefore conditions both the policy action and its
training objective.

\newpage
\paragraph{Real-robot deployment.}

\revision{We evaluate the four YCB-object tasks on a 7-DoF Franka Panda with a
parallel-jaw gripper or 12-DoF XHand, using decoded trajectories as control
references. {All four tasks per end effector are deployed (eight settings), without filtering by simulation success.} Following RoboDojo~\citep{chen2026robodojo}, each task--end-effector
setting undergoes 10 trials with predefined initial layouts replayed before
each trial. Success requires task completion within 1.5 times the
90th-percentile demonstration duration; timeouts and safety-triggered
terminations count as failures.}

\section{Benchmark Distribution}
\label{sec:supp-benchmark}
\label{sec:supp-benchmark-composition}

Each of the 300 benchmark cases pairs a human-video clip with a target robot
embodiment. Figure~\ref{fig:supp-benchmark-distribution} summarizes their
distribution by source dataset, target embodiment, camera setup, and task family.

\begin{figure*}[!t]
\centering
\includegraphics[width=\textwidth]{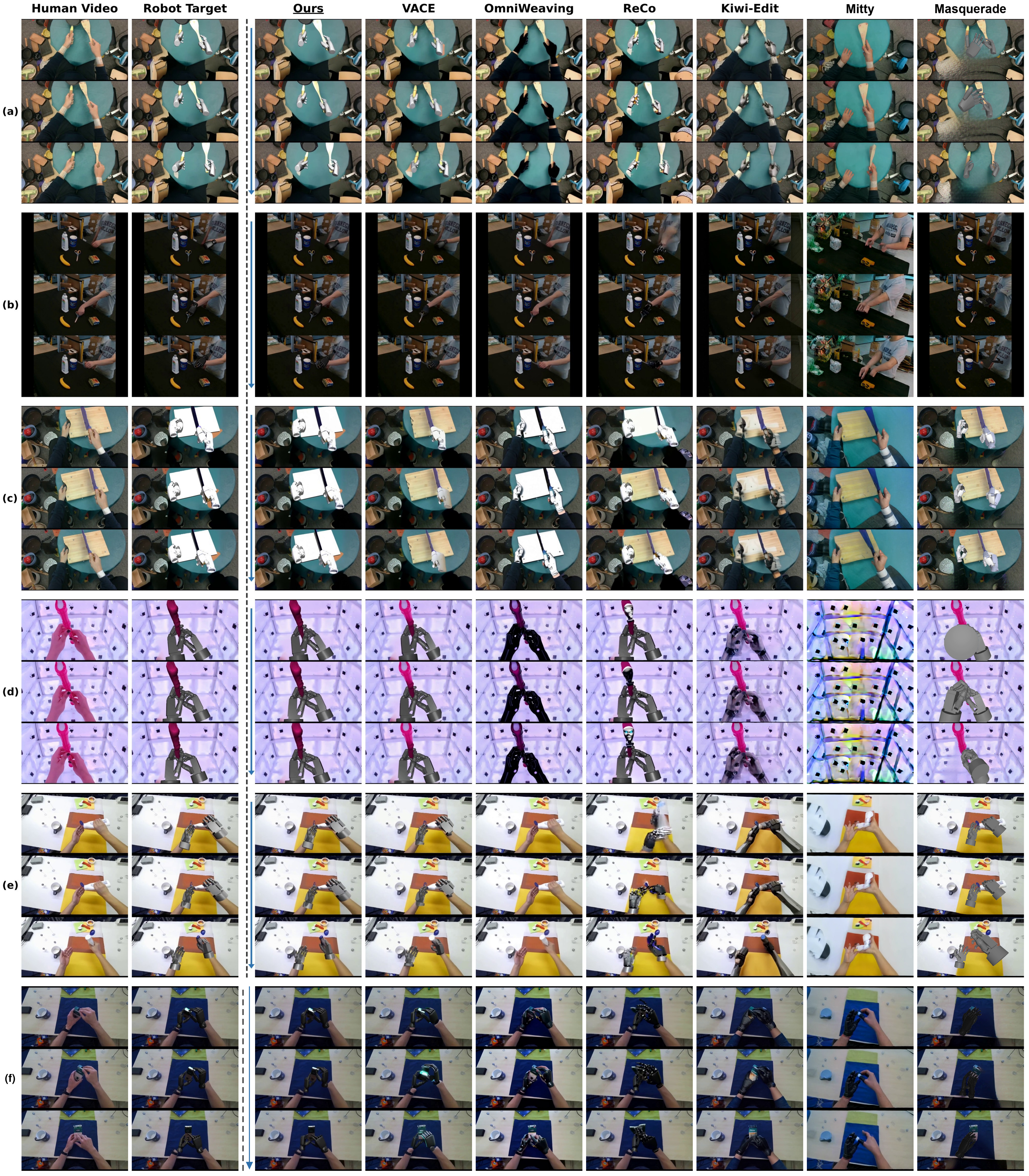}
\caption{Additional qualitative comparisons on six human-object interactions. All methods show the same frames and spatial crops.
}
\label{fig:supp-qualitative}
\end{figure*}

\begin{figure*}[!t]
\centering
\includegraphics[width=\textwidth]{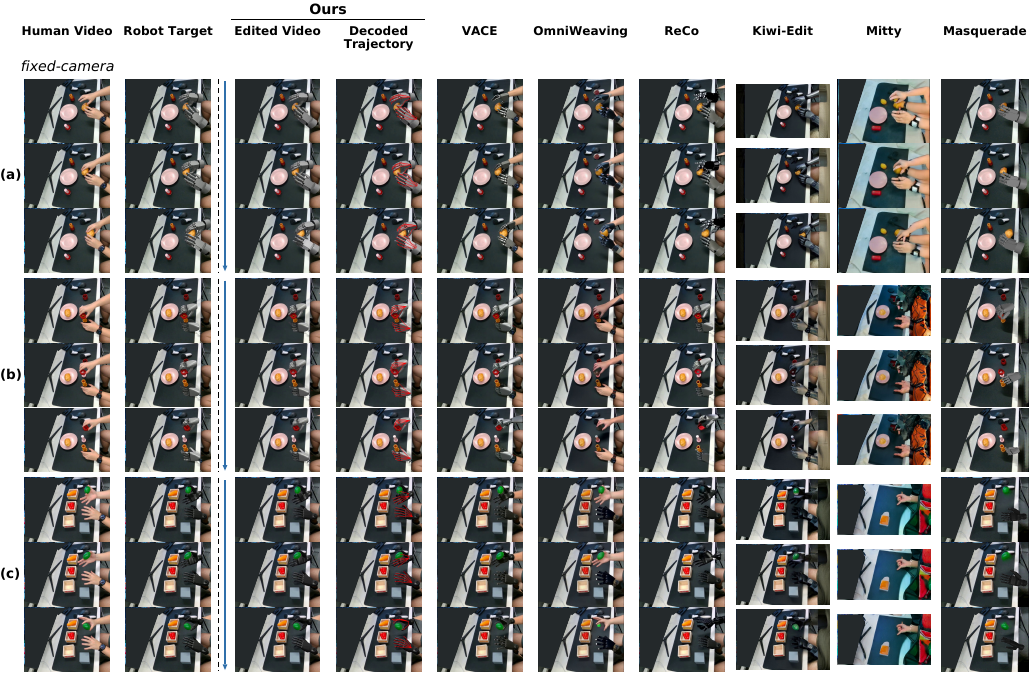}\par
\nointerlineskip
\includegraphics[width=\textwidth]{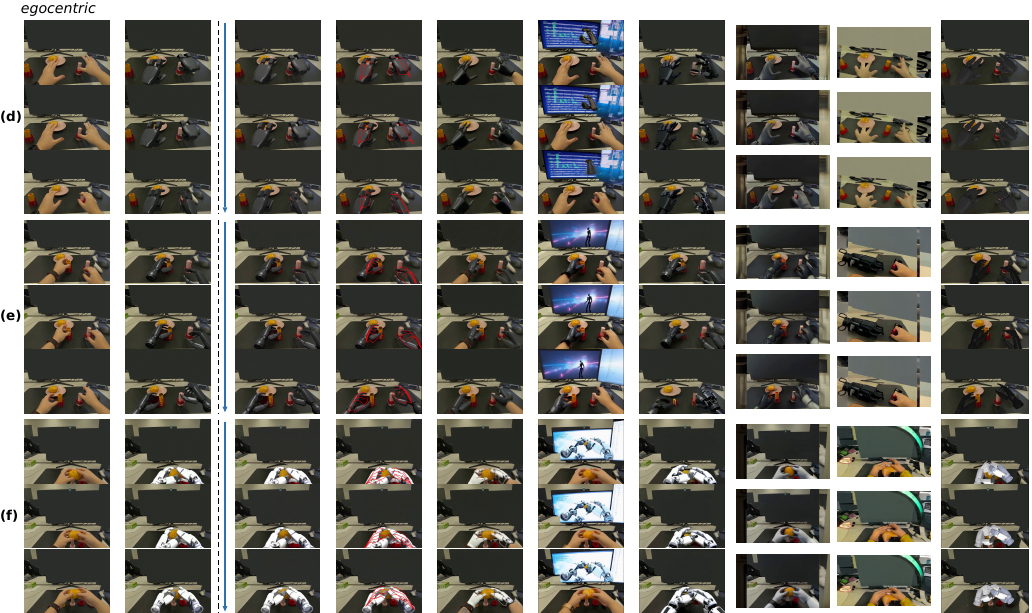}
\caption{{Fixed-camera (a--c) and egocentric (d--f) examples. Our RoboEdit-Trans result is split into edited video and decoded trajectory.}}
\label{fig:in-the-wild}
\end{figure*}

\begin{figure*}[!tp]
    \centering
    \includegraphics[width=0.78\linewidth]{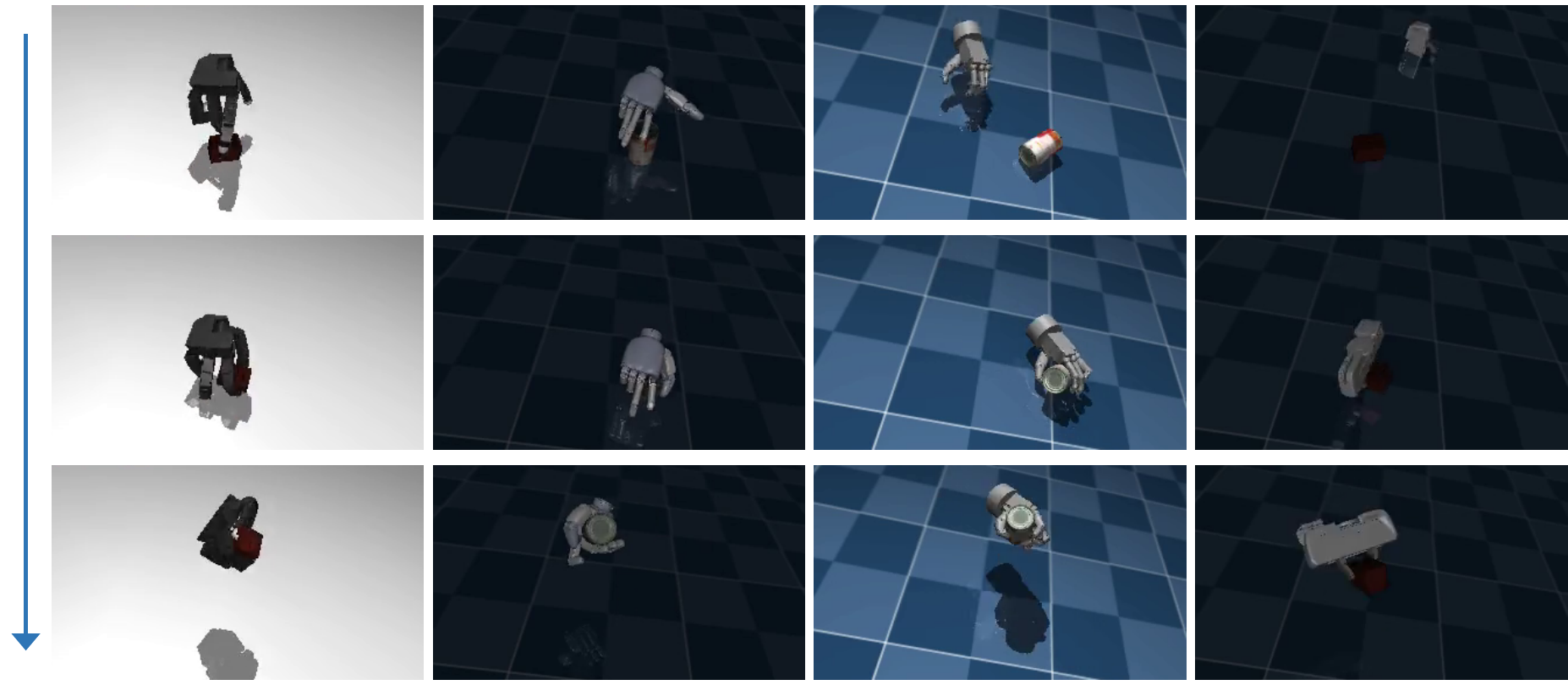}
\caption{Simulation rollouts of the trajectory-conditioned
    controller across four YCB-object manipulation tasks.
}
    \label{fig:supp-simulation-rollouts}

\par\vspace{6pt}
\centering
\includegraphics[width=0.78\linewidth]{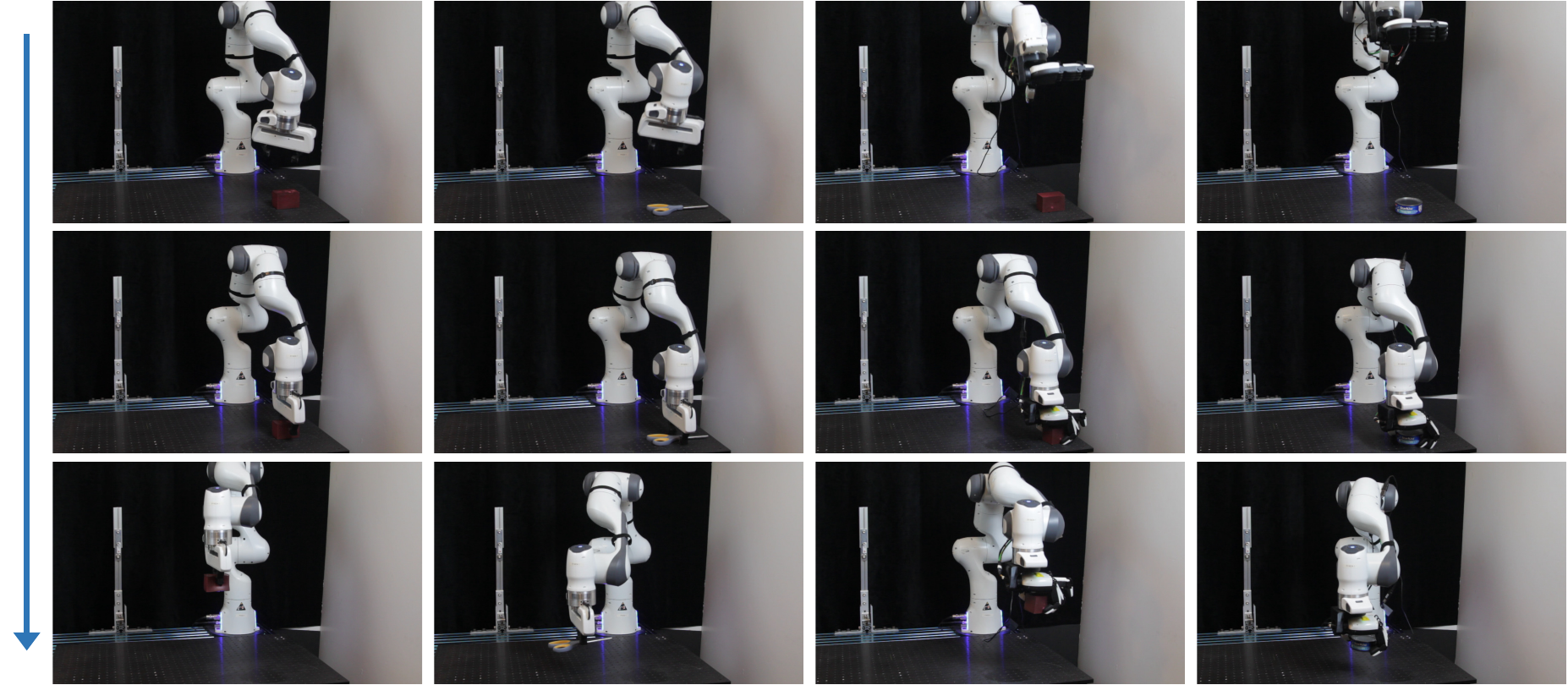}
\caption{Additional real-robot deployment results across YCB-object tasks.
}
\label{fig:supp-rollouts}

\par\vspace{6pt}
\centering
\includegraphics[width=4.6in]{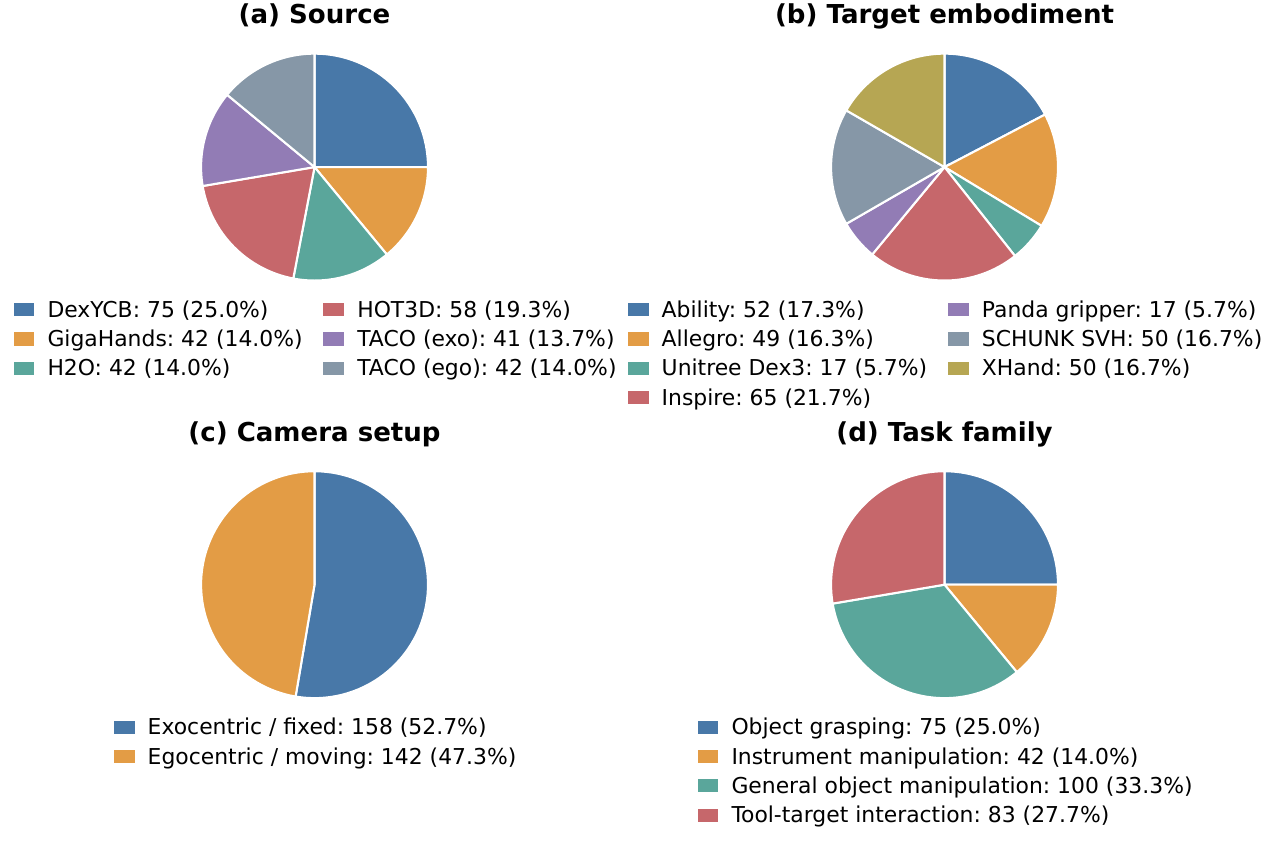}
\caption{Distribution of the 300-case benchmark. Legends show
case counts and percentages; task families are coarse, source-based activity groups.}
\label{fig:supp-benchmark-distribution}
\end{figure*}

\endgroup

\end{document}